%% file: arxiv_mainsupp_v2.tex
\documentclass[letterpaper]{article} 
\usepackage{aaai23}  
\usepackage{times}  
\usepackage{helvet}  
\usepackage{courier}  
\usepackage[hyphens]{url}  
\usepackage{graphicx} 
\usepackage{natbib}  
\usepackage{caption} 
\usepackage{booktabs}
\usepackage{multirow}
\usepackage{amsfonts}
\input{preamble}
\usepackage{newfloat}
\usepackage{listings}
\DeclareCaptionStyle{ruled}{labelfont=normalfont,labelsep=colon,strut=off} 
\floatstyle{ruled}
\newfloat{listing}{tb}{lst}{}
\floatname{listing}{Listing}
\newcommand\underlinecloser[1]{\underline{\smash{#1}}}

\title{RobustLoc: Robust Camera Pose Regression in Challenging Driving Environments}

\author {
    Sijie Wang\textsuperscript{\rm 1}\equalcontrib,
    Qiyu Kang\textsuperscript{\rm 1}\equalcontrib,
    Rui She\textsuperscript{\rm 1}\equalcontrib,
    Wee Peng Tay\textsuperscript{\rm 1}, \\
    Andreas Hartmannsgruber\textsuperscript{\rm 2},
    Diego Navarro Navarro\textsuperscript{\rm 2}
}
\affiliations {
    \textsuperscript{\rm 1} Continental-NTU Corporate Lab, Nanyang Technological University \\
    \textsuperscript{\rm 2} Continental Automotive Singapore \\
    \{wang1679@e.; qiyu.kang@; rui.she@; wptay@\}ntu.edu.sg, \\
    \{andreas.hartmannsgruber; diego.navarro.navarro\}@continental.com
    
}

\usepackage{bibentry}

\begin{document}

\maketitle

\begin{abstract}
Camera relocalization has various applications in autonomous driving. Previous camera pose regression models consider only ideal scenarios where there is little environmental perturbation. To deal with challenging driving environments that may have changing seasons, weather, illumination, and the presence of unstable objects, we propose RobustLoc, which derives its robustness against perturbations from neural differential equations. Our model uses a convolutional neural network to extract feature maps from multi-view images, a robust neural differential equation diffusion block module to diffuse information interactively, and a branched pose decoder with multi-layer training to estimate the vehicle poses. Experiments demonstrate that RobustLoc surpasses current state-of-the-art camera pose regression models and achieves robust performance in various environments. Our code is released at: \url{https://github.com/sijieaaa/RobustLoc}
\end{abstract}

\section{Introduction}\label{sec:introduction}

Accurate camera relocalization plays an important role in autonomous driving.  Given query images, the camera relocalization task aims at estimating the camera poses for which the images are taken. In recent years, many camera relocalization approaches have been proposed. Generally speaking, they fall into two categories.
\begin{enumerate}
    \item One solution is to treat relocalization as a matching task. This solution assumes the availability of a database or a map that stores prior information (e.g., 3D point clouds, images, or descriptors) of sample points. Given a query image, the matching model finds the best match between the query and the database based on a similarity score. The estimated camera pose is then inferred from the matched prior information in the database.
    \item Another solution does not assume the availability of a database and uses only neural networks to regress the camera poses of the query images. This approach constructs an implicit relation between images and poses, which is called \emph{camera pose regression} (CPR).
\end{enumerate}
The prerequisite of a database on the one hand can boost the accuracy of the camera relocalization by storing useful prior information. On the other hand, the computation and storage requirements are proportionate to the number of sample points in the database. To decouple relocalization from the need for a database, there has been a recent surge of research interest in the second category CPR.

\begin{figure}[!t]
\begin{center}
\includegraphics[width=0.43\textwidth]{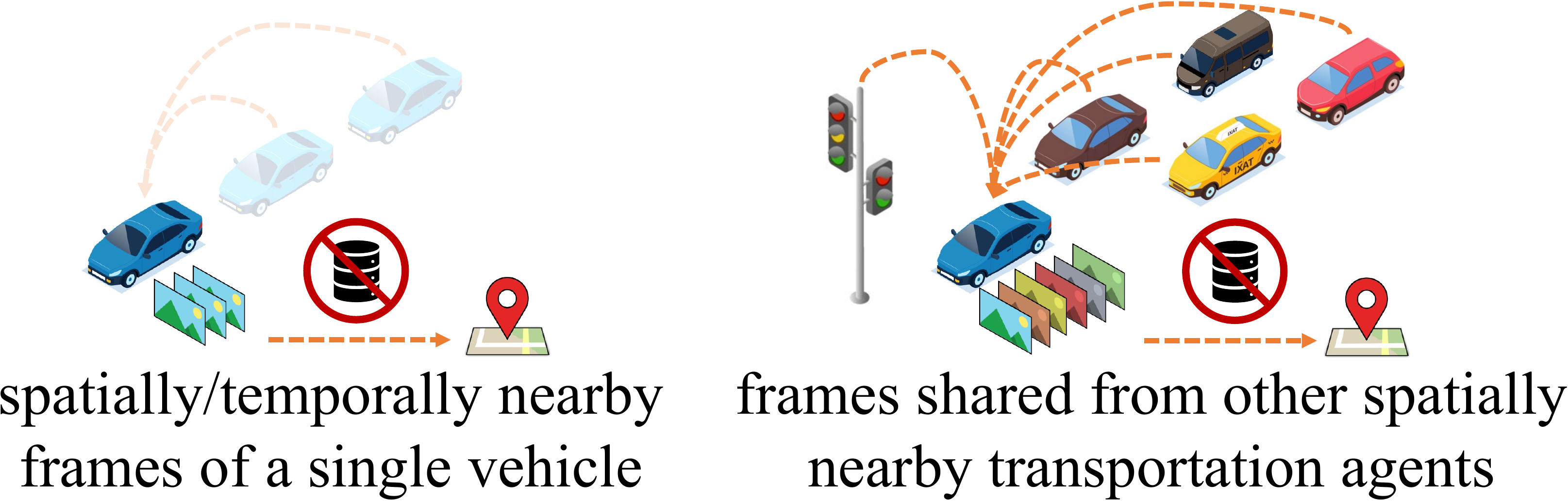}
\end{center}
\caption{Multi-view camera pose regression with neighboring information, without the need for any database. }
\label{fig:neighboring}
\end{figure}

The pioneering work PoseNet \cite{posenet} uses a convolutional neural network (CNN) to extract features from a single image as vector embeddings, and the embeddings are directly regressed to the 6-DoF poses. To further improve the regression performance in driving scenarios, multi-view-based models extend the input from a single image to multi-view images. MapNet \cite{mapnet} leverages pre-computed visual odometry to post-process the output pose trajectory. GNNMapNet \cite{gnnmapnet} integrates a graph neural network (GNN) into CNN to make image nodes interact with neighbors.

The above-mentioned multi-view-based models show promising performance in benign driving environments. To operate well in challenging environments, the model must be robust to environmental perturbations (e.g., changing seasons, weather, illumination, and unstable objects), and effectively leverage neighboring information from spatially or temporally nearby frames of a single vehicle or multi-view images shared from other spatially nearby agents (e.g., using V2X communication) as shown in \cref{fig:neighboring}. Images sharing such neighboring information are said to be \emph{covisible}.

Recently, neural Ordinary Differential Equations (ODEs)  \cite{chen2018neural}  and Partial Differential Equations (PDEs) \cite{chamberlain2021grand, chamberlain2021blend}  have demonstrated their robustness against input perturbations  \cite{yan2019robustness,kang2021Neurips}. Moreover, GNNs can effectively aggregate neighborhood information. We thus propose RobustLoc that not only explores the relations between graph neighbors but also utilizes neural differential equations to improve robustness. We test our new multi-view-based model on three challenging autonomous driving datasets and verify that it outperforms existing state-of-the-art (SOTA) CPR methods.

Our main contributions are summarized as follows:
\begin{enumerate}

\item 
We represent the features extracted from a CNN in a graph and apply graph neural diffusion layers at each stage. I.e., we design feature diffusion blocks at both the feature map extraction and vector embedding stages to achieve robust feature representations. Each diffusion block consists of not only cross-diffusion from node to node in a graph but also self-diffusion within each node. We also propose multi-level training with the branched decoder to better regress the target poses.


\item We conduct experiments in both ideal and challenging noisy autonomous driving datasets to demonstrate the robustness of our proposed method. The experiments verify that our method achieves better performance than the current SOTA CPR methods. 

\item We conduct extensive ablation studies to provide insights into the effectiveness of our design. 
\end{enumerate}


\section{Related Work}

\subsection{Camera Pose Regression}
Given the query images, CPR models directly regress the camera poses of these images without the need for a database. Thus, it does not depend on the scale of the database, which is definitely a born gift compared with those database methods.  \\
PoseNet \cite{posenet} and GeoPoseNet \cite{geoposenet2017} propose the simultaneous learning for location and orientation by integrating balance parameters. MapNet \cite{mapnet} uses visual odometry to serve as the post-processing technique to optimize the regressed poses. LsG \cite{lsg} and LSTM-PoseNet \cite{lstmpose} integrates the sequential information by fusing PoseNet and LSTM. AD-PoseNet and AD-MapNet \cite{adposenet} leverages the semantic masks to drop out the dynamic area in the image. AtLoc \cite{atloc} introduces the global attention to guide the network to learn better representation. GNNMapNet \cite{gnnmapnet} expands the feature exploration from a single image to multi-view images using GNN. IRPNet \cite{irpnet} proposes to use two branches to regress translation and orientation respectively. Coordinet \cite{coordinet} uses the coordconv \cite{coordconv} and weighted average pooling \cite{fc4} to capture spatial relations. There are also some models focusing on LiDAR-based pose regression including PointLoc\cite{wang2021pointloc} and HypLiLoc\cite{wang2023hypliloc}.

\subsection{Neural Differential Equations and Robustness}

The dynamics of a system are usually described by ordinary or partial differential equations. The paper \cite{chen2018neural} first proposes trainable neural ODEs by parameterizing the continuous dynamics of hidden units. The hidden state of the ODE network is modeled as:
\begin{align}
\ddfrac{\mby(t)}{t}=f_{\theta}(\mby(t)) \label{eq:ode_f}
\end{align}
where  $\mby(t)$ denotes the latent state of the trainable network $f_{\theta}$ that is parameterized by weights $\theta$. Recent studies \cite{yan2019robustness,kang2021Neurips} have demonstrated that neural ODEs are intrinsically more robust against input perturbations compared to vanilla CNNs. 


In addition, neural PDEs \cite{chamberlain2021grand, chamberlain2021blend} have been proposed and applied to GNN, where the diffusion process is modeled on the graph. Furthermore, the stability of the heat semigroup and the heat kernel under perturbations of the Laplace operator (i.e., local perturbation of the manifold) is studied \cite{SonKanWan:C22}.

\section{Proposed Model}

In this section, we provide a detailed description of our proposed CPR approach.
We assume that the input is a set of images $\{\mbI_{i}\}_{i\in [N]}$ that may be covisible (see \cref{fig:neighboring}).\footnote{\emph{Notations:} In this paper, we use $[N]$ to denote the set of integers $\{1, 2, \ldots, N\}$.We use boldfaced lowercase letters like $\mbm$ to denote vectors and boldface capital letters like $\mbW$ to denote matrices.} Our objective is to perform CPR on the input images.

\begin{figure*}[ht]
\begin{center}
\includegraphics[width=0.69\textwidth]{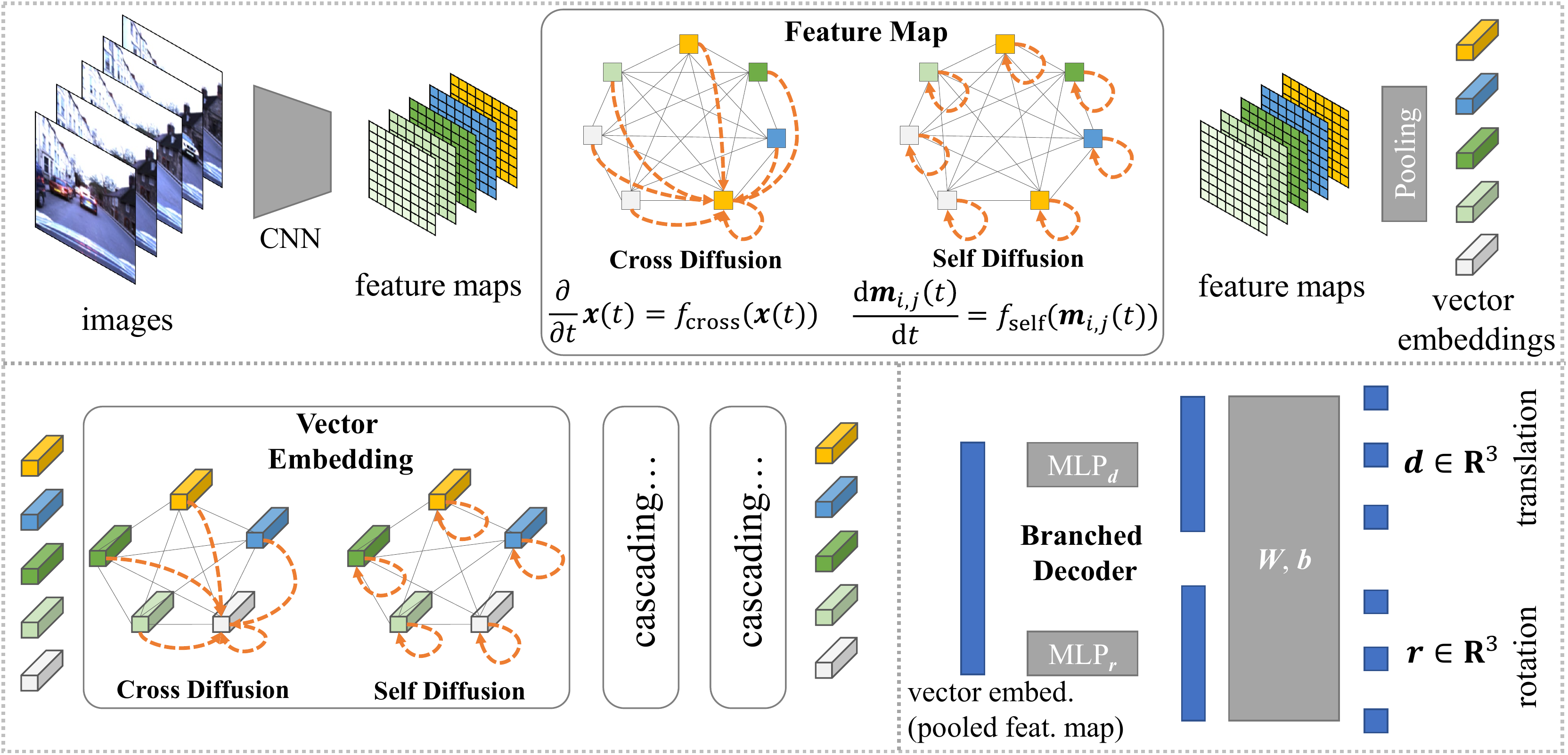}
\end{center}
\caption{The main architecture of RobustLoc. Feature diffusion is performed at both the feature map stage and the vector embedding stage. The branched decoder regresses the 6-DoF poses based on the vector embeddings or the pooled feature maps. The details for multi-layer decoding are shown in \cref{fig:multi-level}. }
\label{fig:model}
\end{figure*}

\subsection{RobustLoc Overview}
We first summarize our multi-view CPR pipeline, which can be decomposed into three different stages, as follows (see \cref{fig:model} and \cref{fig:multi-level}): 
\begin{enumerate}
\item Given $N$ neighboring images, a CNN extracts the feature maps of all these images. Our proposed feature map diffusion block then performs cross-self diffusion on the feature maps.
\item After feature map diffusion, a global average pooling module aggregates the feature maps as vector embeddings, which contain global representations of these images. Similarly, those vector embeddings are then diffused by cascaded diffusion blocks. 
\item Based on the vector embeddings, the branched decoder module regresses the output camera poses. During training, decoding is performed on multiple levels to provide better feature constraints.
\end{enumerate}


\subsection{Neural Diffusion for Feature Maps}
 The input images $\{\mbI_{i}\}_{i\in [N]}$ are passed through a CNN to obtain the feature maps $ \{\mbm_{i} \in \mathbb{R}^{H \times W \times C } \}_{i\in [N]}$. Here, $C$ is the channel dimension, while $H$ and $W$ are the dimensions of a feature map. For each feature map $\mbm_i$, we denote its $j$-th element as $\mbm_{i,j}\in \mathbb{R}^{C} , j\in [HW]$. We next describe the feature map diffusion block, where we perform cross-diffusion from node to node in a graph, and self-diffusion within each node. The two diffusion processes update the feature map by leveraging the neighboring information or only using each node's individual information, respectively.
\subsubsection{Cross-Diffusion Dynamics.}
To support the cross-diffusion over feature maps, we formulate the first graph in our pipeline as:
\begin{align}
\mathcal{G}^{\mathrm{feat}}=(\mathcal{V}^{\mathrm{feat}}, \mathcal{E}^{\mathrm{feat}}), 
\end{align}
where the node set $\mathcal{V}^{\mathrm{\mathrm{feat}}}=\{\mbm_{i,j}\}_{(i,j)\in [N]\times[HW]}$ contains element-wise features $\mbm_{i,j}$ and the edge set $\mathcal{E}^{\mathrm{\mathrm{feat}}}$ is defined as the complete graph edges associated with attention weights as discussed below. And the complete graph architecture is demonstrated to be an effective design shown in \cref{tab:graph}.

To achieve robust feature interaction, we next define the cross-diffusion process as: 
\begin{align}
\frac{\partial}{\partial t} \mbx(t)
& = f_{\mathrm{cross}}(\mbx(t)), \label{eq:pde}
\end{align}
where $f_{\mathrm{cross}}(\mbx(t))$ is a neural network and can be approximately viewed as a neural PDE  with the partial differential operations over a manifold space replaced by the attention modules that we will introduce later. We denote the input to the feature map diffusion module as the initial state at $t=t_{0}$ as $\mbx(t_{0})=\left\{\mbm_{i,j}\right\}_{(i,j)\in [N]\times[HW]}$, where $\mbx(t)=\left\{\mbm_{i,j}(t)\right\}_{(i,j)\in [N]\times[HW]}$ denotes the hidden state of the diffusion.
The diffusion process is known to have robustness against local perturbations of the manifold \cite{chen1998stability} where the local perturbations in our CPR task include challenging weather conditions, dynamic street objects, and unexpected image noise. Therefore, we expect our module \cref{eq:pde} is simultaneously capable of leveraging the neighboring image information and holding robustness against local perturbations.


We next introduce the computation of attention weights in $f_{\mathrm{cross}}(\mbx(t))$ for node features at time $t$. 
We first generate the embedding of each node using multi-head fully connected (FC) layers with learnable parameter matrix $\mbW_{k}$ and bias $\mbb_{k}$ at each head $k=[K]$, where $K$ is the number of heads. The output at each head $k$ can be written as:
\begin{align}
\mbm_{i,j;k}^{\mathrm{FC}}(t)=\mbW_{k} \mbm_{i,j}(t) + \mbb_{k}.
\end{align}
The attention weights are then generated by computing the dot product among all the neighboring nodes using the features 
$\bigl\{\mbm_{i,j;k}^{\mathrm{FC}}(t)\bigr\}_{(i,j)\in[N]\times[HW]}$. We have
\begin{align}
& \{a_{(i,j),(i',j');k}(t)\}_{(i',j')\in \mathcal{N}_{i,j}} \nn
& =\mathrm{Softmax}_{(i',j') \in \mathcal{N}(i,j)}(\mbm_{i,j;k}^{\mathrm{FC}}(t) \cdot  \mbm_{i', j';k}^{\mathrm{FC}}(t)),
\end{align}
where $\mathcal{N}_{i,j}$ denotes the set of neighbors of node $\mbm_{i,j}$.  
Let
\begin{align}
\mbm_{i,j;k}^{\mathrm{weighted}}(t)= \sum_{(i',j') \in \mathcal{N}_{i,j}} a_{(i,j),(i',j');k}(t) \mbm_{i',j';k}^{\mathrm{FC}}(t).
\end{align} Finally, the updated node features are obtained by concatenating the weighted node features from all heads as 
\begin{align}
f_{\mathrm{cross}}(\mbx(t))=\set*{ \concat_{k\in[K]}(\mbm_{i,j;k}^{\mathrm{weighted}}(t)) }_{(i,j)\in [N]\times[HW]}.
\end{align}

Based on the above pipeline, the output of the cross-diffusion at time $t=t_{1}$ can be obtained as:

\begin{align}
\mbx(t_{1})=F_{\mathrm{cross}}(\mbx(t_{0})),
\end{align}
where $F_{\mathrm{cross}}(\cdot)$ denotes the solution of \cref{eq:pde} integrated from $t=t_{0}$ to $t=t_{1}$. 

\subsubsection{Self-Diffusion Dynamics.} In the next step, we update each node feature independently. The node-wise feature update can be regarded as a rewiring of the complete graph to an edgeless graph, and the node-wise feature update is described as:
\begin{align}
\ddfrac{\mbm_{i,j}(t)}{t}=f_{\mathrm{self}}(\mbm_{i,j}(t))=\mathrm{MLP}(\mbm_{i,j}(t)).
\label{eq:NODE}
\end{align}%
And the output of self-diffusion can be obtained as:
\begin{align}
\mbm_{i,j}(t_{2})=F_{\mathrm{self}}(\mbm_{i,j}(t_{1})).
\end{align}
where $F_{\mathrm{self}}(\cdot)$ denotes the solution of \cref{eq:NODE} integrated from $t=t_{1}$ to $t=t_{2}$. 
As neural ODEs are robust against input perturbations  \cite{yan2019robustness,kang2021Neurips}, we expect the updating of each node feature according to the self-diffusion \cref{eq:NODE} to be robust against perturbations like challenging weather conditions, dynamic street objects, and image noise.

\subsection{Vector Embeddings and Diffusion}
After the feature map neural diffusion, we feed the updated feature maps into a global average pooling module to generate the vector embeddings $\{ \mbh_{i}\in \mathbb{R}^{C} \}_{i\in [N]}$, where
\begin{align}
\mbh_{i}=\mathrm{Pooling} (\mbm_{i}) .
\end{align}
Each vector embedding contains rich global representations for the input image together with the information diffused from the neighboring images. To enable diffusion for the global information, we propose to design the vector embedding graph as:
\begin{align}
\mathcal{G}^{\mathrm{vect}}=(\mathcal{V}^{\mathrm{vect}}, \mathcal{E}^{\mathrm{vect}}),  
\end{align}
where the node set $\mathcal{V}^{\mathrm{vect}}=\{\mbh_{i}\}_{i\in[N]}$ contains image vector embeddings $\mbh_{i}$ and the edge set $\mathcal{E}^{\mathrm{vect}}$ is also defined to be the complete graph. Based on this graph $\mathcal{G}^{\mathrm{vect}}$, we construct the cascaded diffusion blocks, to perform global information diffusion. Within the cascaded blocks, each basic diffusion block consists of two diffusion layers: a cross-diffusion layer and a self-diffusion layer, similar to the two diffusion schemes introduced at the feature map diffusion phase.

\subsection{Pose Decoding}
In this subsection, we explain the pose decoding operations.

\subsubsection{Branched Pose Decoder.}\label{decoder}

Each camera pose $\mbp=\{\mbd,\mbr\}\in\mathbb{R}^{6}$, consists of a 3-dimensional translation $ \mbd\in\mathbb{R}^{3}$ and a 3-dimensional rotation $ \mbr\in \mathbb{R}^{3}$. Thus CPR can be viewed as a multi-task learning problem. However, since the translation and rotation elements of $\mbp$ do not scale compatibly, the regression converges in different basins. To deal with it, previous methods consider regression for translation and rotation respectively and demonstrate it is an effective way to improve performance \cite{irpnet}. In our paper, we also follow this insight to design the decoder.

Firstly, the feature embeddings $\{ \mbh_{\mbd},  \mbh_{\mbr}\}$ for translation and rotation are extracted from the feature embedding $\mbh$ using different non-linear MLP layers as:

\begin{align}
\mbh_{\mbd}=\mathrm{MLP}_{\mbd}(\mbh),
\end{align}
\begin{align}
\mbh_{\mbr}=\mathrm{MLP}_{\mbr}(\mbh),
\end{align}
Thus, the features of translation and rotation are decoupled. Next in the second stage, the pose output can be regressed as:

\begin{align}
\mbp=\mbW (\mbh_{\mbd}\concat\mbh_{\mbr}) + \mbb
\end{align}
where $\mbW,\mbb$ are learnable parameters. During training, we compute the regression loss of decoded poses from multiple levels, which we will introduce below. During inference, we use the decoded pose from the last layer as the final output pose.

\subsubsection{Multi-level Pose Decoding Graph.}  
\begin{figure}[!t]
\begin{center}
\includegraphics[width=0.44\textwidth]{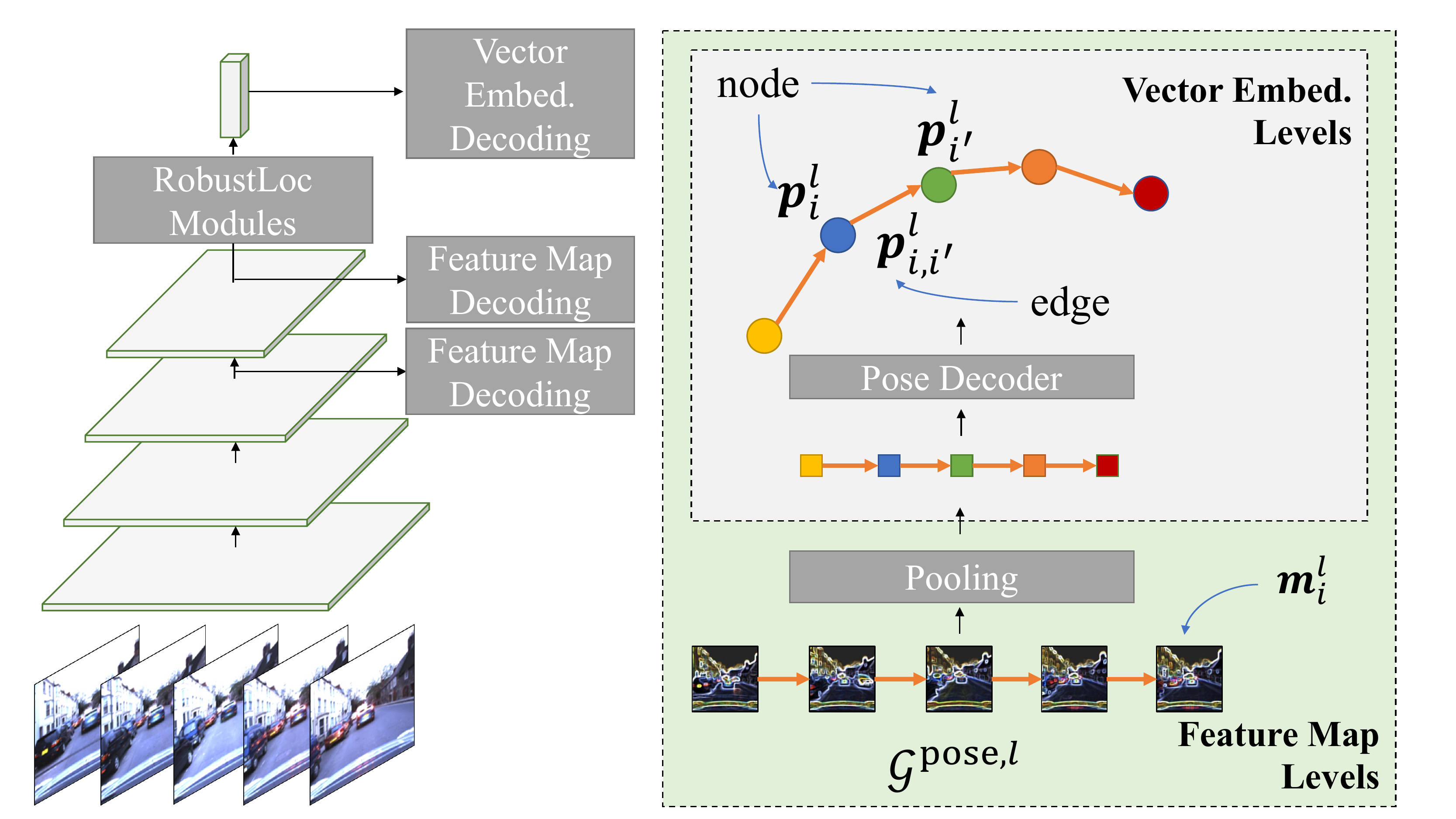}
\end{center}
\caption{Multi-level pose decoding. Decoding can be directly applied to vector embeddings. Feature maps are first pooled and then decoded.  }
\label{fig:multi-level}
\end{figure}

To better regularize the whole regression pipeline, we propose to leverage the feature maps at multiple levels. As shown in \cref{fig:multi-level}, at the vector embedding stage, we use the vector embeddings to regress the poses, while at the feature map stage, we use the feature maps. Denoting the feature maps at layer $l$ as $\{ \mbm_{i}^{l} \in \mathbb{R}^{H \times W \times C} \}_{i\in[N]}$, the pose decoding graph at layer $l$ can be formulated as: 
\begin{align}
\mathcal{G}^{\mathrm{pose},l}=(\mathcal{V}^{\mathrm{pose},l},\mathcal{E}^{\mathrm{pose},l}),
\end{align}
where edge set $\mathcal{E}^{\mathrm{pose},l}$ is defined to be connected with two spatially adjacent nodes which can be viewed as the odometry connection, while the node set $\mathcal{V}^{\mathrm{pose},l}$ is defined depending on layers since the information used to regress poses is different:
\begin{align}
\mathcal{V}^{\mathrm{pose},l}=
\left\{
\begin{array}{ll}
\{ \mbh_{i} \}_{i\in[N]} & \text{if}\  l=L, \\
\{ \mbm_{i}^{l} \}_{i\in[N]} & \text{otherwise}, \\
\end{array}
\right.
\end{align}
where $L$ represents the last layer in our network.

At the last layer where there are vector embeddings, we can directly apply the pose decoder to generate absolute pose messages. By contrast, at feature map layers,  we first apply a global average pooling module on the feature maps to formulate feature vectors, and pose messages can be obtained using the  pose decoder:
\begin{align}
\mbp^{l}_{i}=
\left\{
\begin{array}{l l}
f_{\mathrm{decoder}}^{l}( \mbh^{l}_{i}) & \text{if}\  l=L, \\
f_{\mathrm{decoder}}^{l}( \mathrm{Pooling}(\mbm^{l}_{i})) & \text{otherwise}. \\
\end{array}
\right.
\end{align}
where $f_{\mathrm{decoder}}^{l}(\cdot)$ is the pose decoder at layer $l$. Using the simplified relative pose computation technique in \cite{atloc}, the relative pose messages $\mbp^{l}_{i,i'}$ at layer $l$ can be generated as:
\begin{align}
\mbp^{l}_{i,i'} = \mbp^{l}_{i'} - \mbp^{l}_{i}.
\end{align}
By leveraging multi-layer information, we expect not only the last layer but also the preceding middle-level layers can directly learn the implicit relation between images and poses, which helps to improve the robustness against perturbations.

\begin{table*}[!htp]\footnotesize
\centering
\begin{tabular}{c | l | c  c   c  c     c  c  } 
\toprule
& \multirow{3}{*}{Model} & \multicolumn{2}{c}{Loop (cross-day)} & \multicolumn{2}{c}{Loop (within-day)} & \multicolumn{2}{c}{Full} \\
& & \multicolumn{1}{c}{Mean} & \multicolumn{1}{c}{Median} & \multicolumn{1}{c}{Mean} & \multicolumn{1}{c}{Median} & \multicolumn{1}{c}{Mean} & \multicolumn{1}{c}{Median} \\ 
\midrule
\multirow{5}{*}{\rotatebox{90}{+ Extra Data}}
& GNNMapNet + \emph{post.}  & 7.96 / \underlinecloser{2.56}  & - & - & - & 17.35 / \underlinecloser{3.47} & -  \\

& ADPoseNet & - & - & - & 6.40 / 3.09 & - & 33.82 / 6.77 \\

& ADMapNet & - & - & - & 6.45 / 2.98 & - & 19.18 / 4.60 \\

& MapNet+  & 8.17 / 2.62 & - & - & - & 30.3 / 7.8 &  \\

& MapNet+ + \emph{post.} & 6.73 / \textbf{2.23} & - & - & - & 29.5 / 7.8 &  -\\

\midrule
\multirow{7}{*}{\rotatebox{90}{CPR Only}}
& GeoPoseNet &  27.05 / 18.54 & 6.34 / 2.06  & -  & -  & 125.6 / 27.1 &  107.6 / 22.5 \\

& MapNet & 9.30 / 3.71 &  5.35 / \underlinecloser{1.61} & - & -  & 41.4 / 12.5 &  17.94 / 6.68\\

& LsG & 9.08 / 3.43   & -  & -  & -  & 31.65 / 4.51  &  - \\

& AtLoc &   8.74 / 4.63 & 5.37 / 2.12 & -  & -  & 29.6 / 12.4  & 11.1 / 5.28 \\

& AtLoc+ & \underlinecloser{7.53} / 3.61 &  \underlinecloser{4.06} / 1.98 & -  & -  &  21.0 / 6.15 & 6.40 / 1.50\\

& CoordiNet & -  & -  & \underlinecloser{4.06}  /  \underline{1.44} & \underline{2.42}  /  \underline{0.88} &  \underline{14.96}  /  5.74  & \textbf{3.55}  /  \underline{1.14} \\

& RobustLoc (ours) & \textbf{4.68} / 2.67  & \textbf{3.70} / \textbf{1.50} & \textbf{2.49} /  \textbf{1.40}  & \textbf{1.97} / \textbf{0.84} & \textbf{9.37} / \textbf{2.47}  & \underline{5.93} / \textbf{1.06} \\
\bottomrule
\end{tabular}
\caption{Median and mean translation/rotation estimation error (m/$^\circ$) on the Oxford RobotCar dataset. The best and the second-best results in each metric are highlighted with bold and underline respectively. ``-'' denotes no data provided.
}
\label{tab:robotcar}
\end{table*}

\subsection{Loss Function}
Following the approach in \cite{atloc}, we use a weighted balance loss for translation and rotation predictions. For the input image $I_{i}$, we denote the translation and rotation targets as $\mbd_{i}^{*} \in \mathbb{R}^{3}$ and $\mbr_{i}^{*} \in \mathbb{R}^{3}$ respectively. Then the absolute pose loss term $\mathcal{L}^{l}_{i}$ and the relative pose loss term $\mathcal{L}^{l}_{i,i^{'}}$ at decoding layer $l$ are computed as: 
\begin{align}
&\ml{\mathcal{L}^{l}_{i} = \norm{\mbd_{i}^{l}-\mbd_{i}^{*}} \exp(-\alpha) + \alpha \\
+ \norm{\mbr_{i}^{l}-\mbr_{i}^{*}} \exp(-\beta) + \beta,}\\
&\ml{\mathcal{L}^{l}_{i,i^{'}} = \norm{\mbd_{i,i^{'}}^{l}-\mbd_{i,i^{'}}^{*}} \exp(-\gamma) + \gamma \\ 
+ \norm{\mbr_{i,i^{'}}^{l}-\mbr_{i,i^{'}}^{*}} \exp(-\lambda) + \lambda},
\end{align}
where $\mbd_{i}^{l},\mbr_{i}^{l},\mbd_{i,i^{'}}^{l},\mbr_{i,i^{'}}^{l}$ are outputs at layer $l$, while $\alpha, \beta, \gamma, \lambda$ are all learnable parameters. Finally, the overall loss function can be obtained as:
\begin{align}
\mathcal{L}=\sum_{l\in\{3,4,L\}}\sum_{i
\in[N], i^{'}\in \mathcal{N}^l_i}\mathcal{L}^{l}_{i} + \mathcal{L}^{l}_{i,i^{'}},
\end{align}
where $\calN^l_i$ is the neighborhood of node $i$ in $\calG^{\mathrm{pose},l}$. We use the logarithmic form of the quaternion to represent rotation $\mbr$ as:
\begin{align}
\mbr=\log \mbq=\left\{
\begin{array}{ll}
\frac{(\mbq_{2},\mbq_{3},\mbq_{4})}{\norm{(\mbq_{2},\mbq_{3},\mbq_{4})}}\cos^{-1} \mbq_{1} &\ \text{if}\  \norm{(\mbq_{2},\mbq_{3},\mbq_{4})} \not= 0,\\
0 &\ \text{otherwise},
\end{array}
\right.
\end{align}
where $\mbq = (\mbq_{1}, \mbq_{2}, \mbq_{3}, \mbq_{4}) \in \mathbb{R}^{4}$ represents a quaternion.

\section{Experiments}
In this section, we first evaluate our proposed model on three large autonomous driving datasets. We next present an ablation study to demonstrate the effectiveness of our model design.

\subsection{Datasets and Implemention Details}
\subsubsection{Oxford RobotCar.}
The Oxford RobotCar dataset\cite{robotcar} is a large autonomous driving dataset collected by a car driving along a route in Oxford, UK. It consists of two different routes: 1) Loop with a trajectory area of $8.8\times10^{4}\mathrm{m}^{2}$ and length of $10^{3}\mathrm{m}$,  and 2) Full with a trajectory area of $1.2\times10^{6}\mathrm{m}^{2}$ and length of $9\times10^{3}\mathrm{m}$.

\subsubsection{4Seasons.}
There are only a few existing methods designed for robust CPR in driving environments, and the experiment on the Oxford dataset is insufficient for comparison. Thus we also conduct experiments on another driving dataset to cover more driving scenarios. The 4Seasons dataset \cite{4seasons} is a comprehensive dataset for autonomous driving SLAM. It was collected in Munich, Germany, covering varying perceptual conditions. Specifically, it contains different environments including the business area, the residential area, and the town area. In addition, it consists of a wide variety of weather conditions and illuminations. In our experiments, we use 1) Business Campus (business area), 2) Neighborhood (residential area), and 3) Old Town (town area).

\subsubsection{Perturbed RobotCar.}
\begin{figure}[!t]
\begin{center}
\includegraphics[width=0.32\textwidth]{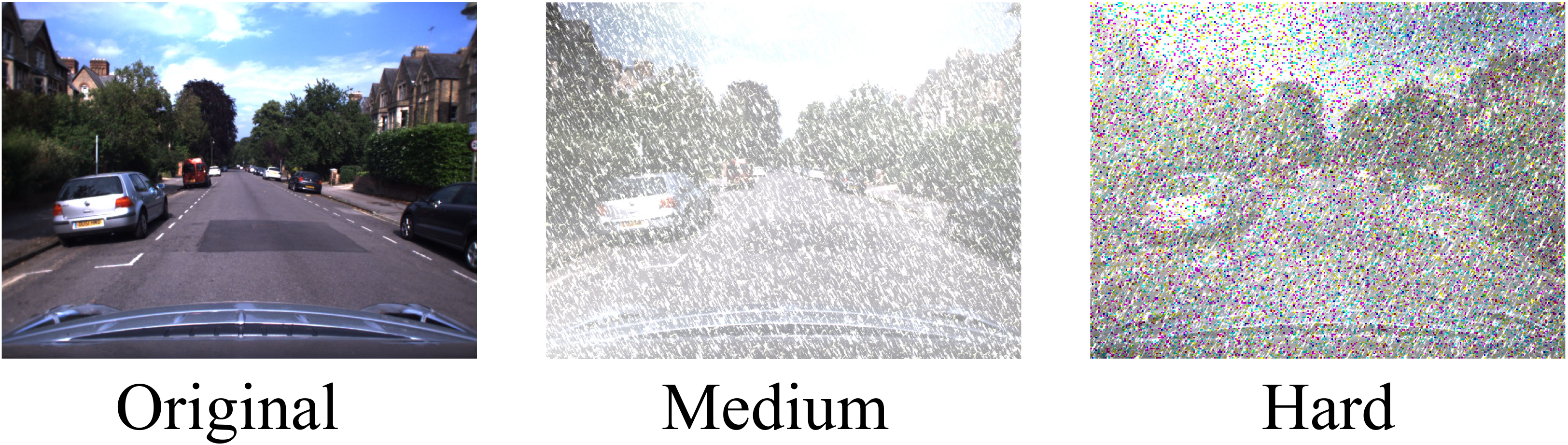}
\end{center}
\caption{Visualization of the Perturbed RobotCar dataset. Medium is with fog, snow, rain, and spatter on the lens. Hard is with added Gaussian noise.}
\label{fig:noisy}
\end{figure}

To further evaluate the performance under challenging environments, we inject noise into the RobotCar Loop dataset and call this the Perturbed RobotCar dataset as shown in \cref{fig:noisy}. We create three scenarios: 1) Medium (with fog, snow, rain, and spatter on the lens), 2) Hard (with added Gaussian noise), and 3) Hard (+ \emph{noisy training}) (i.e., training with noisy augmentation).

\subsubsection{Implemention}
We use ResNet34 as the backbone, which is pre-trained on the ImageNet dataset. We set the maximum number of input images as $11$. We resize the shorter side of each input image to $128$ and set the batch size to $64$. The Adam optimizer with a learning rate $2\times10^{-4}$ and weight decay $5\times10^{-4}$ is used to train the network. Data augmentation techniques include random cropping and color jittering. We set the integration times $t_{0}=0$, $t_{1}=1$, and $t_{2}=2$. The number of attention heads is 8. We train our network for $300$ epochs. All of the experiments are conducted on an NVIDIA A5000. 


\subsection{Main Results}\label{subsec:main results}
On the Oxford RobotCar dataset, as shown in \cref{tab:robotcar}, we obtain the best performance in 10 out of 12 metrics. Using the mean error, which is easily influenced by outlier predictions, RobustLoc outperforms the baselines by a significant margin. In the most challenging route Full, to the best of our knowledge, RobustLoc is the first to achieve less than $10\mathrm{m}$ mean translation error for CPR.

The 4Seasons dataset consists of more varied driving scenes. As shown in \cref{tab:4seasons}, RobustLoc achieves the best performance in 11 out of 12 metrics. Again, using the mean error metric, RobustLoc outperforms the baselines by a significant margin.

On the Perturbed RobotCar dataset, where the images contain more challenging weather conditions and noisy perturbations, RobustLoc achieves the best in all metrics. The superiority of RobustLoc over other baselines is more obvious in \cref{tab:noisy robotcar}.

\begin{table*}[!htp]\footnotesize
\centering
\begin{tabular}{l | c c  c c  c c   } 
\toprule
\multirow{2}{*}{Model} & \multicolumn{2}{c}{Business Campus} & \multicolumn{2}{c}{Neighborhood}  & \multicolumn{2}{c}{Old Town}\\
& \multicolumn{1}{c}{Mean}& \multicolumn{1}{c}{Median}& \multicolumn{1}{c}{Mean}& \multicolumn{1}{c}{Median}& \multicolumn{1}{c}{Mean}& \multicolumn{1}{c}{Median}\\
\midrule
GeoPoseNet  & 11.04 / 5.78 & 5.93 / 2.03 & 2.87 / 1.30 & 1.92 / 0.88   & 64.81 / 6.67 & 15.03 / 1.57  \\

MapNet & 10.35 / 3.78  & 5.66 / 1.83  & 2.81 / 1.05   & 1.89 / 0.92   & 46.56 / 7.14  & 16.52 / 2.12  \\

GNNMapNet & \underlinecloser{7.69} / 4.34 & \underlinecloser{5.52} / 2.16 & 3.02 / 2.92  & 2.14 / 1.45 & \underlinecloser{41.54} / 7.30 & 19.23 / 3.26 \\

AtLoc & 11.53 / 4.84 & 5.81 / \underlinecloser{1.50} & 2.80 / 1.16 & 1.83 / 0.93  & 84.17 / 7.81 & 17.10 / 1.73 \\

AtLoc+   & 13.70 / 6.41 & 5.58 / 1.94 & 2.33 / 1.39 & 1.61 / 0.88 & 68.40 / 5.51 &  14.52 / 1.69 \\

IRPNet &  10.95 / 5.38  & 5.91 / 1.82 & 3.17 / 2.85   & 1.98 / 0.90 & 55.86 / 6.97    &  17.33 / 3.11  \\

CoordiNet  & 11.52 / \underlinecloser{3.44} & 6.44 / \textbf{1.38} & \underlinecloser{1.72} / \underlinecloser{0.86} & \underlinecloser{1.37} / \underlinecloser{0.69} & 43.68 / \underlinecloser{3.58} &  \underlinecloser{11.83} / \underlinecloser{1.36} \\

RobustLoc (ours)  & \textbf{4.28} / \textbf{2.04} & \textbf{2.55} / \underlinecloser{1.50} & \textbf{1.36} / \textbf{0.83} & \textbf{1.00} / \textbf{0.65} & \textbf{21.65} / \textbf{2.41} &  \textbf{5.52} / \textbf{1.05} \\
\bottomrule
\end{tabular}
\caption{Median and mean translation/rotation estimation error (m/$^\circ$) on the 4Seasons dataset. The best and the second-best results in each metric are highlighted with bold and underline respectively.
}
\label{tab:4seasons}
\end{table*}


\begin{table*}[!htp]\footnotesize
\centering
\begin{tabular}{l |  c c c c  c c } 
\toprule

\multirow{2}{*}{Model} & \multicolumn{2}{c}{Medium} & \multicolumn{2}{c}{Hard} & \multicolumn{2}{c}{Hard (+ \emph{noisy training})} \\
& \multicolumn{1}{c}{Mean} & \multicolumn{1}{c}{Median} & \multicolumn{1}{c}{Mean}& \multicolumn{1}{c}{Median} 
& \multicolumn{1}{c}{Mean} & \multicolumn{1}{c}{Median}  \\
\midrule
GeoPoseNet  & 20.47 / 8.76  & 8.70 / 2.30  & 41.71 / 17.63 & 14.02 / 3.13  & 24.03 / 11.14  &  7.14 / 1.70\\
MapNet   & 17.93 / 7.01 & 6.89 / 2.00 & 49.36 / 20.01  & 18.37 / \underlinecloser{2.58}  & 21.22 / 8.38  & 6.38 / 1.97 \\
GNNMapNet  &  \underlinecloser{16.17} / 7.24    &  8.02 / 2.35 &  73.97 / 35.57 &  61.47 / 19.73  & \underlinecloser{14.55} / \underlinecloser{7.62}  & 6.69 / \underlinecloser{1.57} \\

AtLoc  & 19.92 / 7.25 & 7.26 / \underlinecloser{1.74} & 52.56 / 23.46 &  15.01 / 3.17   &  23.48 / 11.43 &  7.42 / 2.38\\
AtLoc+  & 17.68 / 7.48 & \underlinecloser{6.19}  / 1.80 & \underlinecloser{37.92} / 18.65  & \underlinecloser{12.17} / 2.93   & 22.61 / 11.23  & \underlinecloser{6.21} / 1.83 \\
IRPNet  & 16.35 / 7.56 & 8.71 / 2.28    & 45.72 / 21.84  & 17.99 / 3.50   &  24.73 / 11.20 &  6.73 / 1.82\\
CoordiNet   & 17.67 / \underlinecloser{6.66} & 7.63 / 1.79 & 44.11 / \underlinecloser{16.42}  & 17.21 / 2.70   &  24.06 / 12.27 &  6.25 / 1.61 \\
RobustLoc (ours)  & \bf{8.12}  /  \bf{3.83}  & \textbf{5.34} / \textbf{1.53}   & \textbf{27.75} / \textbf{9.70}  & \textbf{11.59} / \textbf{2.64}   & \textbf{10.06} / \textbf{4.95}  & \textbf{5.18} / \textbf{1.43} \\
\bottomrule
\end{tabular}
\caption{Median and mean translation/rotation estimation error (m/$^\circ$) on the Perturbed RobotCar dataset. The best and the second-best results in each metric are highlighted with bold and underline respectively. RobustLoc achieves the best in all metrics.
}
\label{tab:noisy robotcar}
\end{table*}

\subsection{Analysis}
\subsubsection{Ablation Study.}
We justify our design for RobustLoc by ablating each module. From \cref{tab:ablation study}, we observe that every module in our design contributes to the final improved estimation. We see that making use of neighboring information from covisible frames and learning robust feature maps contribute to more accurate CPR.

\begin{table}[!htb]\footnotesize
\centering
\begin{tabular}{l  c  c } 
\toprule
\multirow{1}{*}{Method} & \multicolumn{1}{c}{ Mean Error (m/$^\circ$) on Loop (c.)} \\
\midrule
base model  &  8.38 / 4.29 \\
+ feature map graph & 7.01 / 3.86 \\
+ vector embedding graph & 6.24 / 3.21\\
+ diffusion &  5.53 / 2.95\\
+ branched decoder &  5.14 / 2.79\\
+ multi-level decoding & \textbf{4.68} / \textbf{2.67} \\
\midrule
diffusion at stage 3 & 5.27 / 2.90  \\
diffusion at stage 3,4 & 4.86 / 3.18 \\
diffusion at stage 4 & \textbf{4.68} / \textbf{2.67}  \\
multi-layer concatenation & 5.80 / 3.26  \\
\midrule
more augmentation & \textbf{4.68} / \textbf{2.67}  \\
less augmentation & 5.32 / 3.17  \\
\bottomrule
\end{tabular}
\caption{Ablation study, diffusion design, and augmentation design comparison on the Oxford RobotCar dataset.}
\label{tab:ablation study}
\end{table}

\subsubsection{Salience Visualization.}
\begin{figure}[!t]
\begin{center}
\includegraphics[width=0.3\textwidth]{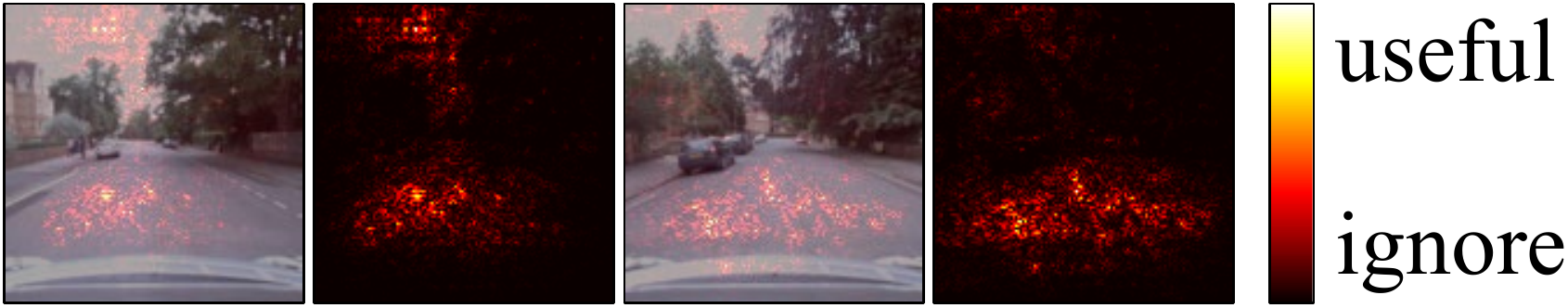}
\end{center}
\caption{Robust features from RobustLoc. }
\label{fig:salience}
\end{figure}

Salience maps shown in \cref{fig:salience} suggest that in driving environments, RobustLoc pays more attention to relatively robust features such as the skyline and the road, similar to PixLoc \cite{pixloc}. In addition, dynamic objects such as vehicles are implicitly suppressed in RobustLoc's regression pipeline.

\subsubsection{Diffusion and Augmentation.}
Using multi-level features is an effective method in dense prediction tasks such as depth estimation \cite{yan2021cadepth}. To test if this holds in CPR, we use the feature maps from the lower stage 3 (see \cref{fig:multi-level}), which however does not lead to performance improvement shown in \cref{tab:ablation study}. We also utilize the multi-level concatenation strategy used in GNNMapNet. This does not lead to significant changes. These experiments demonstrate that CPR benefits more from high-level features with more semantic information than from low-level local texture features. Finally, we test the performance when training with less data augmentation, which leads to worse performance. This suggests that more extensive data augmentation can enhance the model robustness in challenging scenarios, which is consistent with the experimental results on the Perturbed Robotcar dataset in \cref{tab:noisy robotcar}. 


\subsubsection{Graph Design.}
We next explore the use of different graph designs for feature map diffusion and vector embedding diffusion. The grid graph stacks an image with two other spatially adjacent images as a cube, and the attention weights are formulated within the 6-neighbor area (for feature maps) or the 2-neighbor area (for vector embeddings). The self-cross graph computes attention weights first within each image and then across different images. From \cref{tab:graph}, we see that the complete graph has the best performance. This is because, in the complete graph, each node can interact with all other nodes, allowing the aggregation of useful information with appropriate attention weights.

\begin{table}[!htb]\footnotesize
\centering
\begin{tabular}{l  c   c} 
\toprule
\multirow{1}{*}{Method} & \multicolumn{1}{c}{Mean Error (m/$^\circ$) on Full} \\
\midrule
grid graph & 15.67 / 2.95  \\
self-cross graph & 15.31 / 3.28  \\
complete graph & \textbf{9.37} / \textbf{2.47}  \\
\midrule
\multirow{1}{*}{} & \multicolumn{1}{c}{Mean Error ($^\circ$) on Business Campus}  \\
\midrule
quaternion & 2.23  \\
Lie group & 2.20  \\
rotation matrix & 2.25  \\
log (quaternion) & \textbf{2.04}  \\
\bottomrule
\end{tabular}
\caption{Graph design comparison on the Oxford RobotCar dataset and rotation representation comparison on the 4Seasons dataset.}
\label{tab:graph}
\end{table}

\subsubsection{Rotation Representation.}
We compare different representations of rotation in \cref{tab:graph}, where the log form of the quaternion is the optimal choice. The other three representations, including the vanilla quaternion, the Lie group, and the vanilla rotation matrix, show similar performance.


\subsubsection{Trajectory Visualization.}
\begin{figure}[!t]
\begin{center}
\includegraphics[width=0.4\textwidth]{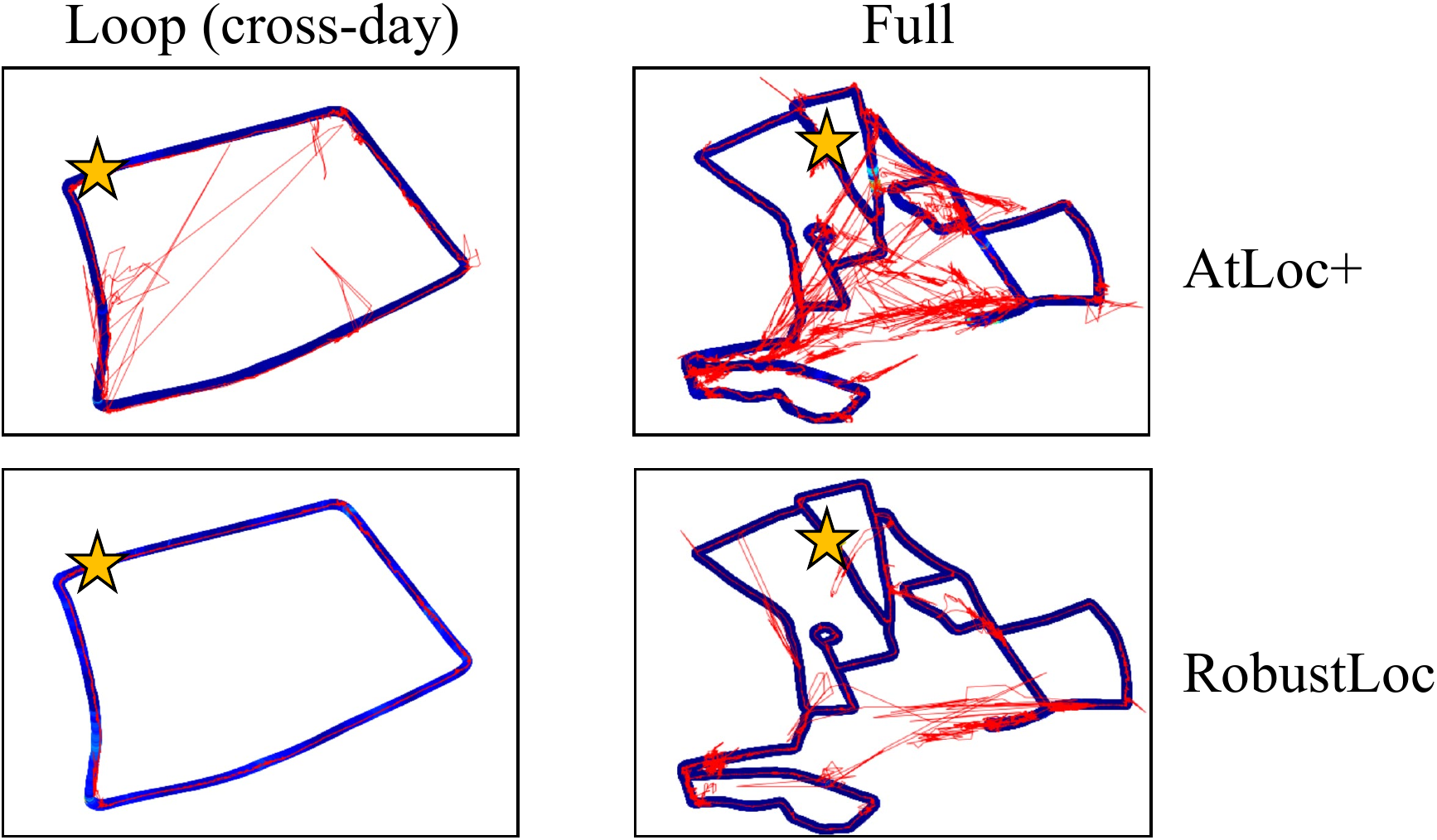}
\end{center}
\caption{Trajectory visualization on the Oxford RobotCar dataset. The ground truth trajectories are shown in bold blue lines, and the estimated trajectories are shown in thin red lines. The stars mark the start of the trajectories.}
\label{fig:trajectory}
\end{figure}

We visualize the output pose trajectories as shown in \cref{fig:trajectory}, where a significant gap can be seen from the comparison. RobostLoc outputs more smooth and globally accurate poses compared with the previous method, which shows the effectiveness of our design.

\subsubsection{Inference Speed.}
We finally test the performance using a different number of input frames. The inference speed does not drop significantly when increasing the input frames. And even the slowest one (using $11$ frames) can run $50$ iterations per second and achieve real-time regression.  On the other hand, more frames can bring performance improvement when the input size is small, while further increasing frame size does not bring significant change.

\begin{table}[!htp]\footnotesize
\centering
\begin{tabular}{l  c   c   c c c} 
\toprule
\multirow{1}{*}{\#frames} & \multicolumn{1}{c}{3} & \multicolumn{1}{c}{5} & \multicolumn{1}{c}{7} & \multicolumn{1}{c}{9} & \multicolumn{1}{c}{11} \\
\midrule
Speed (iters/s)  & \textbf{56}  & 55  & 53  &  52  &  50\\
Mean Error (m)   &  5.28  &  5.09  &  4.96 & \textbf{4.68}   & 4.72 \\
\bottomrule
\end{tabular}
\caption{The performance using different numbers of frames on the Oxford RobotCar Loop (cross-day).}
\vspace{-0.5cm}
\label{tab:additional insight}
\end{table}

\section{Conclusion}
We have proposed and verified the performance of a robust CPR model RobustLoc. The model's robustness derives from the use of information from covisible images and neural graph diffusion to aggregate neighboring information, which is present in challenging driving environments. 

\section{Acknowledgments}
This work is supported under the RIE2020 Industry Alignment Fund-Industry Collaboration Projects (IAF-ICP) Funding Initiative, as well as cash and in-kind contribution from the industry partner(s), and by the National Research Foundation, Singapore and Infocomm Media Development Authority under its Future Communications Research \& Development Programme. The computational work for this article was partially performed on resources of the National Supercomputing Centre, Singapore (\url{https://www.nscc.sg}).


\bibliography{aaai23}

\clearpage

\newpage



\begin{center}
  {
  \large
  \lineskip .5em
  \begin{tabular}[t]{c}
        {\LARGE\bf Supplement \par}
  \end{tabular}
  \par
  }
  \vskip .5em
  \vspace*{12pt}
\end{center}

\begin{figure}[!htb]
\begin{center}
\includegraphics[width=0.47\textwidth]{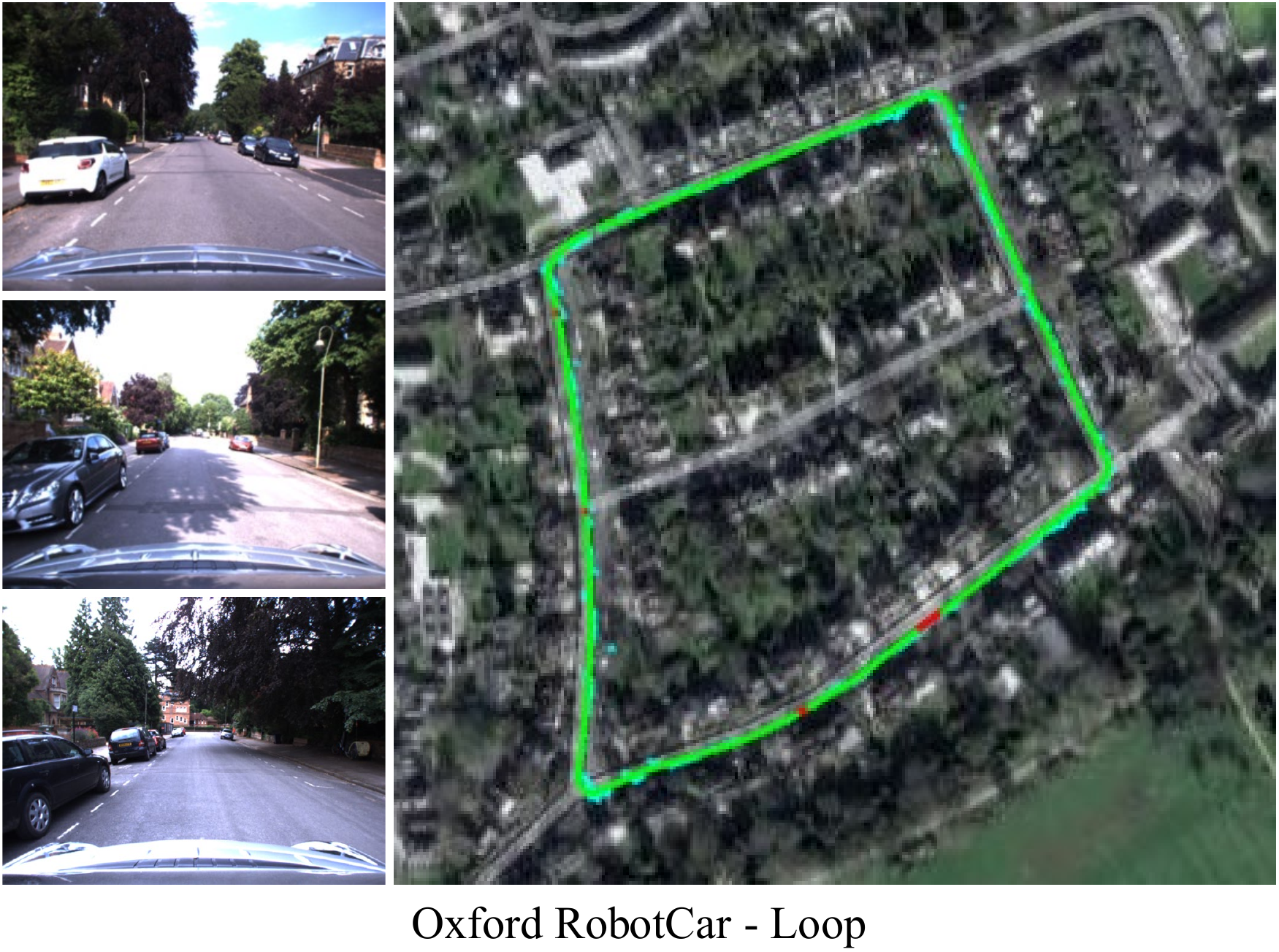}
\end{center}
\caption{Visualization of the Oxford RobotCar Loop dataset. The Loop (cross-day) and the Loop (within-day) share the same images but have different data splits.}
\label{figs:supp_loop}
\end{figure}

\begin{figure}[!htb]
\begin{center}
\includegraphics[width=0.47\textwidth]{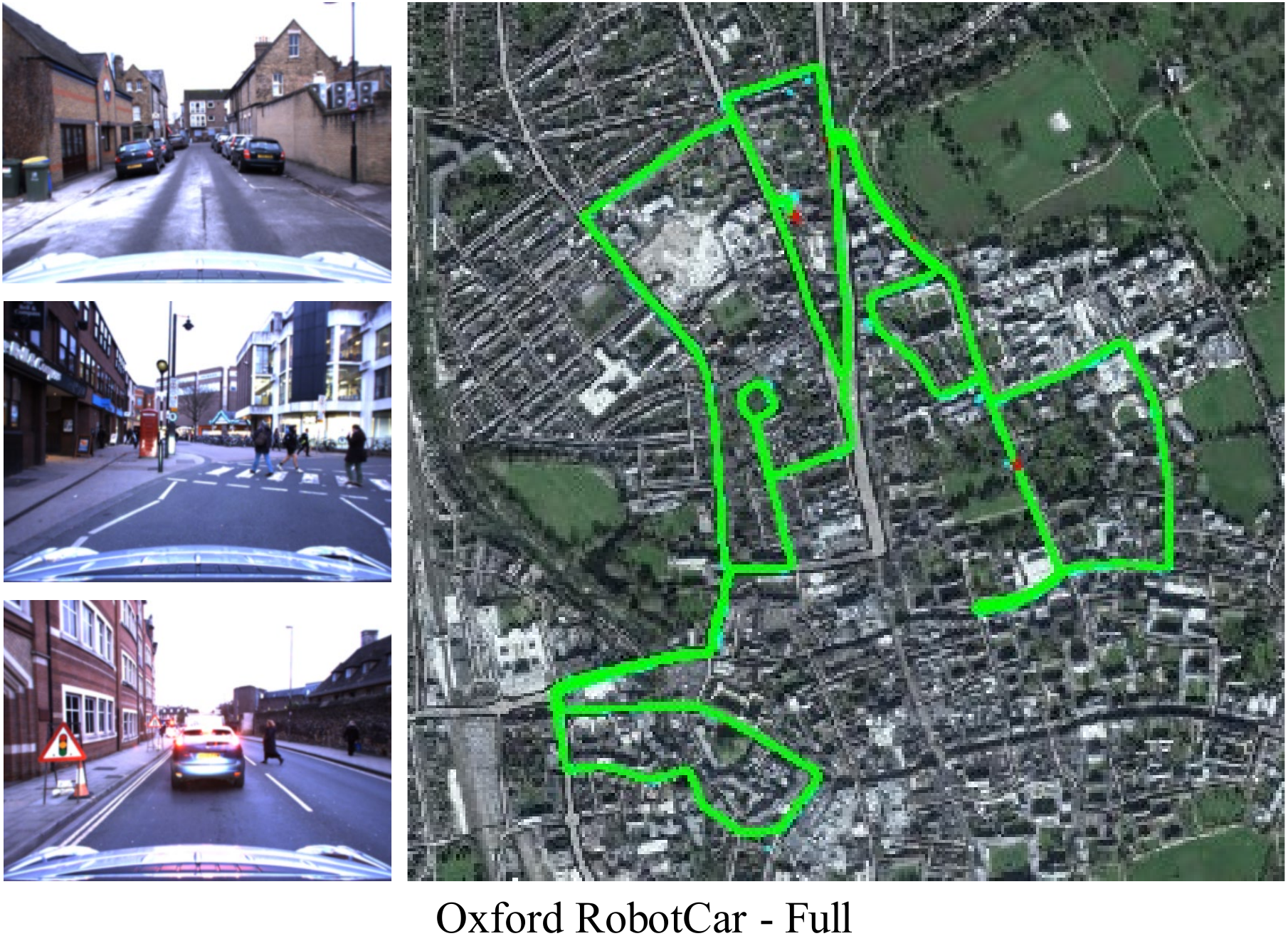}
\end{center}
\caption{Visualization of the Oxford RobotCar Full dataset.}
\label{figs:supp_full}
\end{figure}

\begin{figure}[!htb]
\begin{center}
\includegraphics[width=0.47\textwidth]{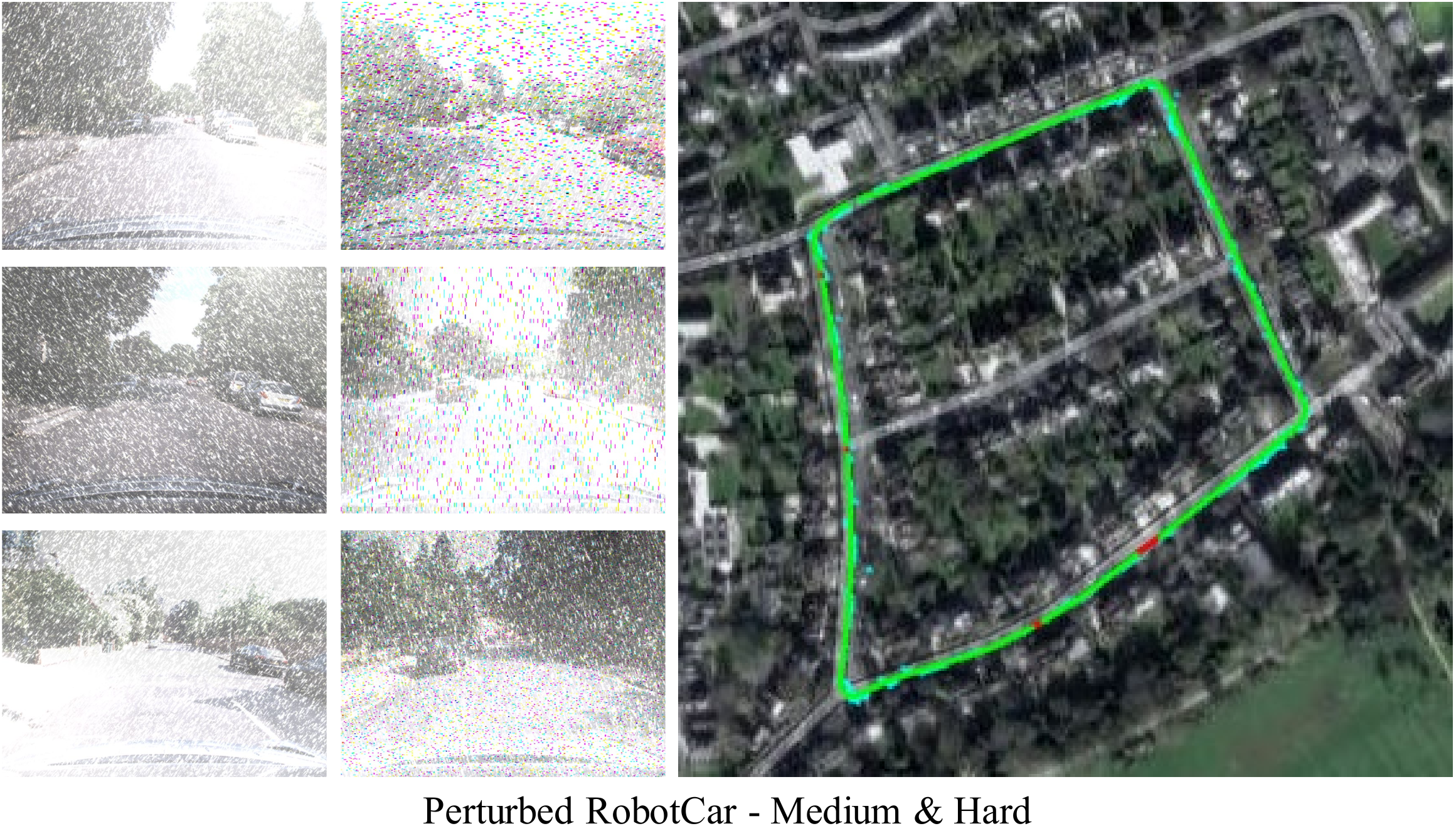}
\end{center}
\caption{Visualization of the Perturbed RobotCar dataset.}
\label{figs:supp_perturbed}
\end{figure}

\begin{figure}[!htb]
\begin{center}
\includegraphics[width=0.47\textwidth]{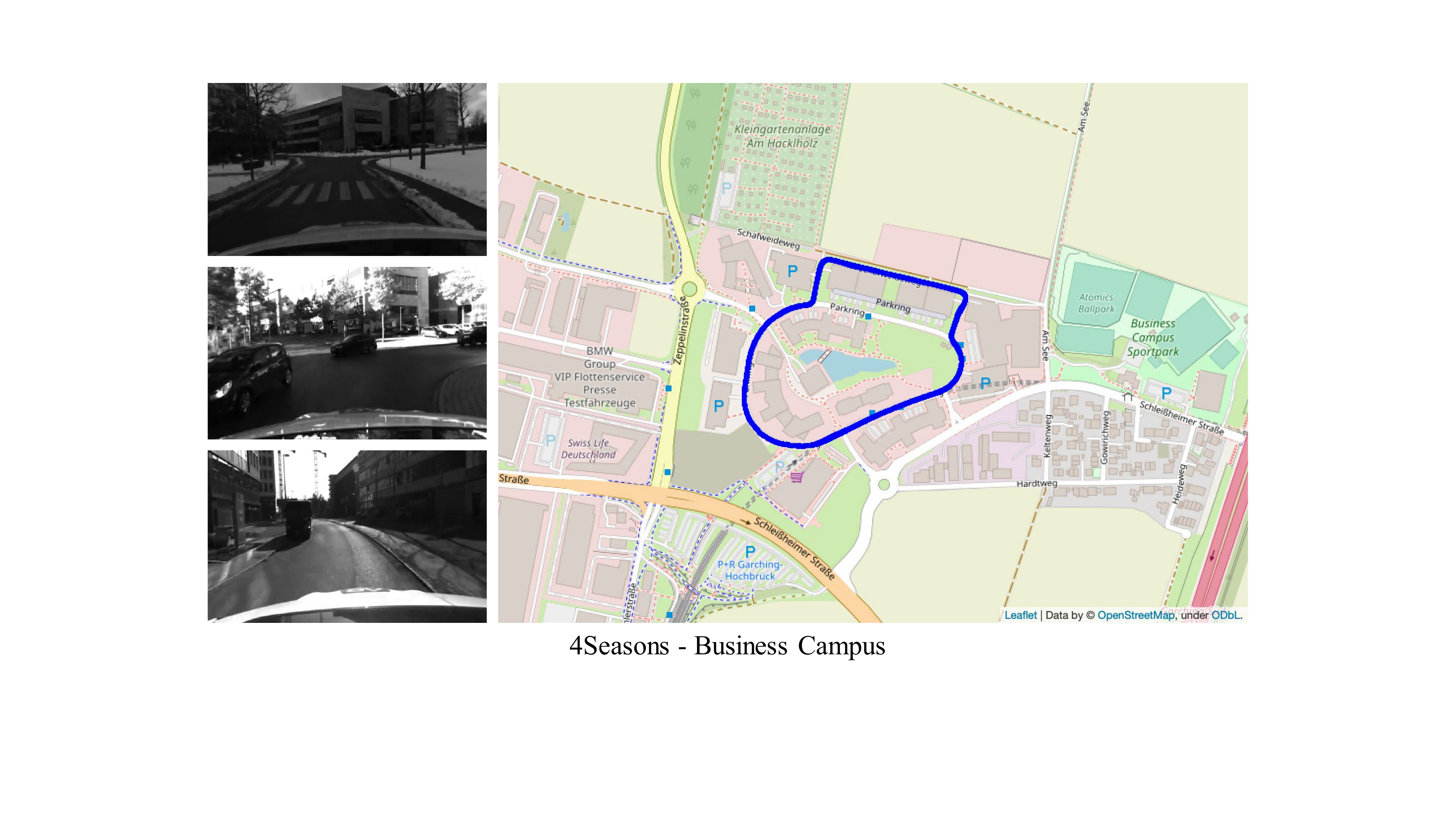}
\end{center}
\caption{Visualization of the 4Seasons Business Campus dataset.}
\label{figs:supp_business}
\end{figure}

\begin{figure}[!htb]
\begin{center}
\includegraphics[width=0.47\textwidth]{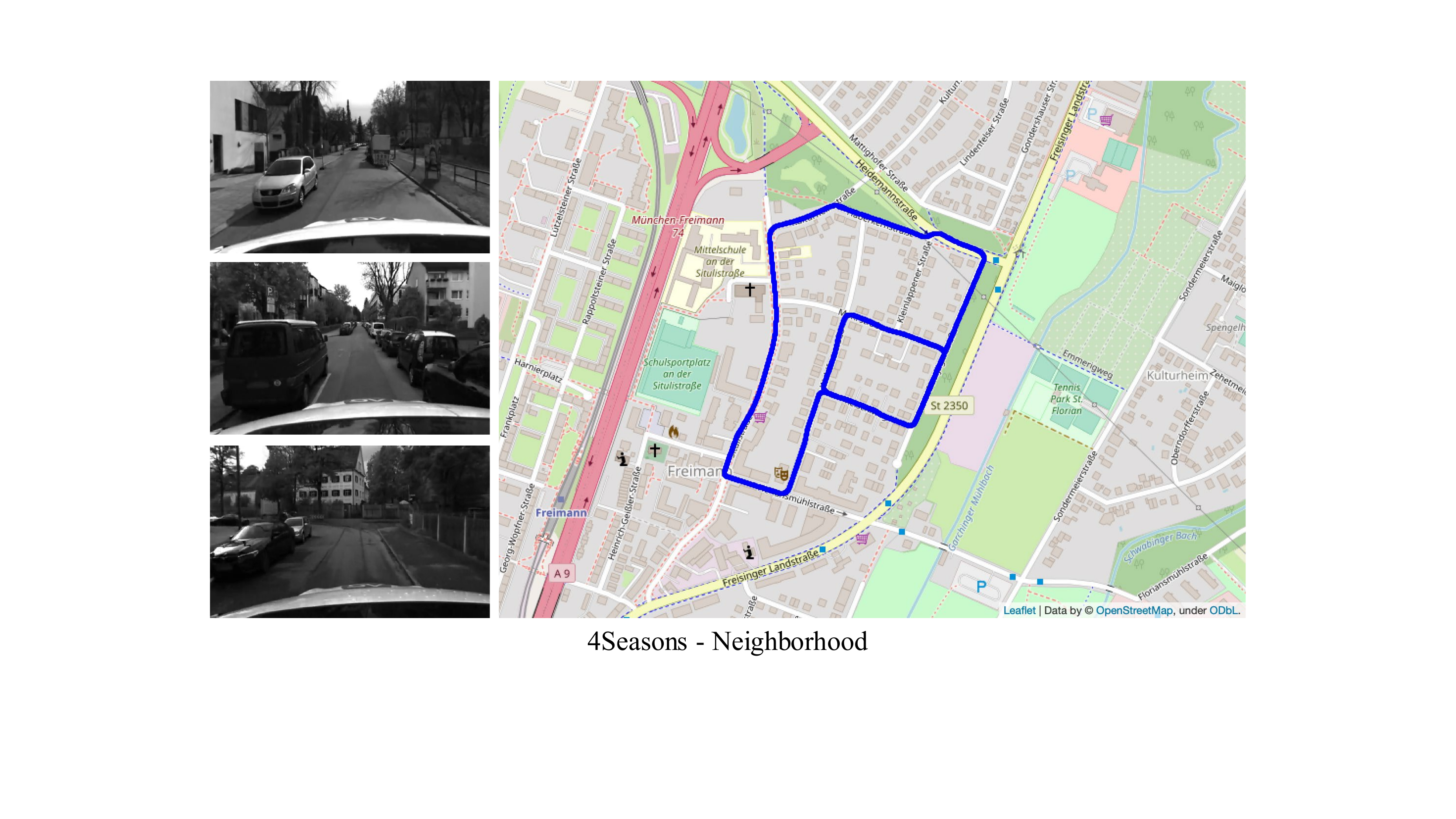}
\end{center}
\caption{Visualization of the 4Seasons Neighborhood dataset.}
\label{figs:supp_neighborhood}
\end{figure}

\begin{figure}[!htb]
\begin{center}
\includegraphics[width=0.47\textwidth]{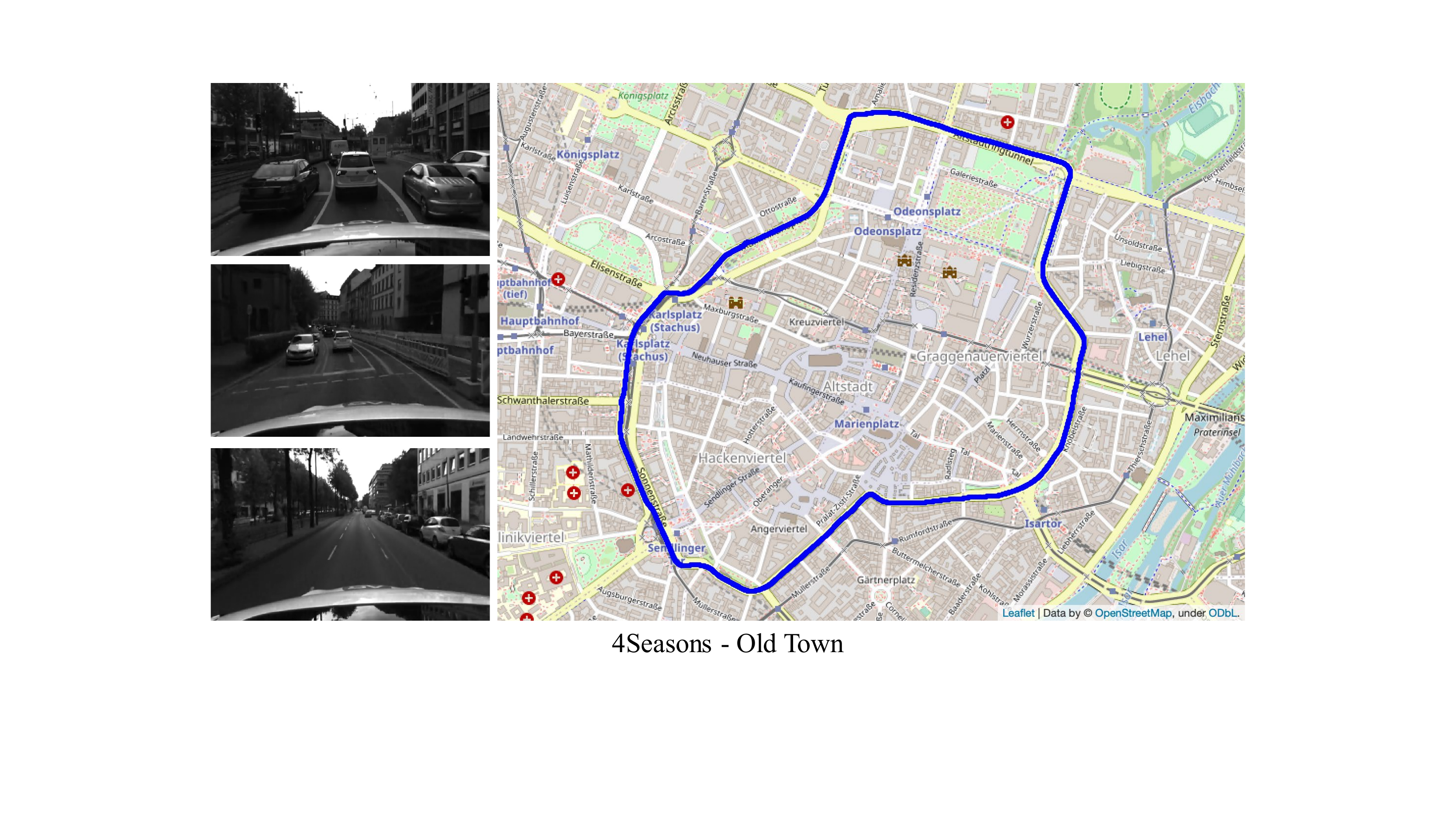}
\end{center}
\caption{Visualization of the 4Seasons Old Town dataset.}
\label{figs:supp_oldtown}
\end{figure}

\section{Baseline Models}
The baseline models in our comparison include: PoseNet, GeoPoseNet, MapNet, LsG, AtLoc/AtLoc+, GNNMapNet, ADPoseNet/ADMapNet, IRPNet, CoordiNet.

\section{Dataset Configuration}
The datasets we used in our experiments include the Oxford Robotcar dataset, the 4Seasons dataset, and the Perturbed Robotcar dataset. All of the datasets are available online at:
\begin{itemize}
\item \url{https://robotcar-dataset.robots.ox.ac.uk/},
\item \url{https://vision.cs.tum.edu/data/datasets/4seasons-dataset/download}.
\end{itemize}
For each dataset and each scene, we list the corresponding data split as shown in  \cref{stab:oxford config}, \cref{stab:4seasons config}, and \cref{stab:perturbed robotcar config}.

\section{Dataset Visualization}
We visualize part of the images from the datasets we used in our experiments as shonw in \cref{figs:supp_loop}, \cref{figs:supp_full}, \cref{figs:supp_perturbed}, \cref{figs:supp_business}, \cref{figs:supp_neighborhood}, and \cref{figs:supp_oldtown}.

\begin{table}[!htb]\footnotesize
\centering
\begin{tabular}{l | c  |  c   } 
\toprule
Dataset & Train & Test \\
\midrule
Loop (cross-day) 
&  2014-06-26-08-53-56  &  2014-06-23-15-36-04 \\
&  2014-06-26-09-24-58  &  2014-06-23-15-41-25 \\
\midrule
Loop (within-day) 
&  2014-06-26-09-24-58  &  2014-06-26-08-53-56 \\
&  2014-06-23-15-41-25  &  2014-06-23-15-36-04 \\
\midrule
Full 
&  2014-11-28-12-07-13  &  2014-12-09-13-21-02 \\
&  2014-12-02-15-30-08  &   \\

\bottomrule
\end{tabular}
\caption{The Oxford RobotCar configuration.}
\label{stab:oxford config}
\end{table}

\begin{table}[!htb]\footnotesize
\centering
\begin{tabular}{l | c  |  c   } 
\toprule
Dataset & Train & Test \\
\midrule
Business Campus
&  2020-10-08--09-30-57  &  2021-02-25--14-16-43\\
&  2021-01-07--13-12-23  &  \\
\midrule
Neighborhood
& 2020-03-26--13-32-55		 & 2021-05-10--18-32-32		  \\
& 2020-10-07--14-47-51		 &  \\
& 2020-10-07--14-53-52		 &  \\
& 2020-12-22--11-54-24		 &  \\
& 2021-02-25--13-25-15		 &  \\
& 2021-05-10--18-02-12		 &  \\
\midrule
Old Town
& 2020-10-08--11-53-41	 &  2021-02-25--12-34-08	 \\
& 2021-01-07--10-49-45	 &   \\
& 2021-05-10--21-32-00	 &   \\
\bottomrule
\end{tabular}
\caption{The 4Seasons configuration.}
\label{stab:4seasons config}
\end{table}

\begin{table}[!htb]\footnotesize
\centering
\begin{tabular}{l | c  |  c   } 
\toprule
Dataset & Train & Test \\
\midrule
Medium
&  2014-06-26-08-53-56  &  2014-06-23-15-36-04 \\
&  2014-06-26-09-24-58  &  2014-06-23-15-41-25 \\
\midrule
Hard
&  2014-06-26-08-53-56  &  2014-06-23-15-36-04 \\
&  2014-06-26-09-24-58  &  2014-06-23-15-41-25 \\
\bottomrule
\end{tabular}
\caption{The Perturbed RobotCar configuration.}
\label{stab:perturbed robotcar config}
\end{table}

\section{Codebase}
Our codes are developed based on the following repositories:
\begin{itemize}
\item \url{https://github.com/BingCS/AtLoc},
\item \url{https://github.com/psh01087/Vid-ODE}.
\end{itemize}
Our codes is released at:\\
\url{https://github.com/sijieaaa/RobustLoc}

\end{document}

%% file: preamble.tex
\usepackage[T1]{fontenc}
\usepackage[utf8]{inputenc}
\usepackage{mathtools}
\usepackage{amssymb,mathrsfs}
\usepackage{amsthm}
\usepackage{bm}
\usepackage{scalerel}
\usepackage{nicefrac}
\usepackage{microtype} 
\usepackage[shortlabels]{enumitem}
\usepackage{graphicx}
\usepackage{epstopdf}
\DeclareGraphicsExtensions{.eps,.png,.jpg,.pdf}

\usepackage{url}
\usepackage{colortbl}
\usepackage{booktabs}
\usepackage{multirow}
\usepackage[table,dvipsnames]{xcolor}
\usepackage[normalem]{ulem}
\usepackage{xparse}
\usepackage{calc}
\usepackage{etoolbox}

\makeatletter
\@ifpackageloaded{natbib}{
	\relax
}{
	\usepackage{cite}
}
\makeatother

\usepackage{array}
\newcolumntype{L}[1]{>{\raggedright\let\newline\\\arraybackslash\hspace{0pt}}m{#1}}
\newcolumntype{C}[1]{>{\centering\let\newline\\\arraybackslash\hspace{0pt}}m{#1}}
\newcolumntype{R}[1]{>{\raggedleft\let\newline\\\arraybackslash\hspace{0pt}}m{#1}}

\makeatletter
\let\MYcaption\@makecaption
\makeatother
\usepackage[font=footnotesize]{subcaption}
\makeatletter
\let\@makecaption\MYcaption
\makeatother

\usepackage{glossaries}
\makeatletter
\sfcode`\.1006

\let\oldgls\gls
\let\oldglspl\glspl

\newcommand\fussy@ifnextchar[3]{%
	\let\reserved@d=#1%
	\def\reserved@a{#2}%
	\def\reserved@b{#3}%
	\futurelet\@let@token\fussy@ifnch}
\def\fussy@ifnch{%
	\ifx\@let@token\reserved@d
		\let\reserved@c\reserved@a
	\else
		\let\reserved@c\reserved@b
	\fi
	\reserved@c}

\renewcommand{\gls}[1]{%
\oldgls{#1}\fussy@ifnextchar.{\@checkperiod}{\@}}
\renewcommand{\glspl}[1]{%
\oldglspl{#1}\fussy@ifnextchar.{\@checkperiod}{\@}}

\newcommand{\@checkperiod}[1]{%
	\ifnum\sfcode`\.=\spacefactor\else#1\fi
}

\robustify{\gls}
\robustify{\glspl}
\makeatother

\newacronym{wrt}{w.r.t.}{with respect to}
\newacronym{RHS}{R.H.S.}{right-hand side}
\newacronym{LHS}{L.H.S.}{left-hand side}
\newacronym{iid}{i.i.d.}{independent and identically distributed}

\usepackage{float}

\ifx\loadbibentry\undefined
	\relax
\else
	\usepackage{bibentry}
\fi

\usepackage[capitalize]{cleveref}
\crefname{equation}{}{}
\Crefname{equation}{}{}
\crefname{claim}{claim}{claims}
\crefname{step}{step}{steps}
\crefname{line}{line}{lines}
\crefname{condition}{condition}{conditions}
\crefname{dmath}{}{}
\crefname{dseries}{}{}
\crefname{dgroup}{}{}

\crefname{Problem}{Problem}{Problems}
\crefformat{Problem}{Problem~(#2#1#3)}
\crefrangeformat{Problem}{Problems~(#3#1#4) to~(#5#2#6)}

\crefname{Theorem}{Theorem}{Theorems}
\crefname{Corollary}{Corollary}{Corollaries}
\crefname{Proposition}{Proposition}{Propositions}
\crefname{Lemma}{Lemma}{Lemmas}
\crefname{Definition}{Definition}{Definitions}
\crefname{Example}{Example}{Examples}
\crefname{Assumption}{Assumption}{Assumptions}
\crefname{Remark}{Remark}{Remarks}
\crefname{Rem}{Remark}{Remarks}
\crefname{remarks}{Remarks}{Remarks}
\crefname{Appendix}{Appendix}{Appendices}
\crefname{Supplement}{Supplement}{Supplements}
\crefname{Exercise}{Exercise}{Exercises}
\crefname{Theorem_A}{Theorem}{Theorems}
\crefname{Corollary_A}{Corollary}{Corollaries}
\crefname{Proposition_A}{Proposition}{Propositions}
\crefname{Lemma_A}{Lemma}{Lemmas}
\crefname{Definition_A}{Definition}{Definitions}

\usepackage{algorithm}
\usepackage{algpseudocode}

\ifx\loadbreqn\undefined
	\relax
\else
	\usepackage{breqn}
\fi

\makeatletter
\def\cleartheorem#1{%
    \expandafter\let\csname#1\endcsname\relax
    \expandafter\let\csname c@#1\endcsname\relax
}
\def\clearthms#1{ \@for\tname:=#1\do{\cleartheorem\tname} }
\makeatother

\ifx\renewtheorem\undefined
	\ifx\useTheoremCounter\undefined
		\newtheorem{Theorem}{Theorem}
		\newtheorem{Corollary}{Corollary}
		\newtheorem{Proposition}{Proposition}
		
	\else

	\fi

\fi

\theoremstyle{remark}

\theoremstyle{plain}

\newcommand{\qednew}{\nobreak \ifvmode \relax \else
		\ifdim\lastskip<1.5em \hskip-\lastskip
			\hskip1.5em plus0em minus0.5em \fi \nobreak
		\vrule height0.75em width0.5em depth0.25em\fi}



\newcommand{\ml}[1]{\begin{multlined}[t]#1\end{multlined}}
\newcommand{\nn}{\nonumber\\ }

\NewDocumentCommand{\movedownsub}{e{^_}}{%
	\IfNoValueTF{#1}{%
		\IfNoValueF{#2}{^{}}
	}{%
		^{#1}
	}%
	\IfNoValueF{#2}{_{#2}}
}

\let\latexchi\chi
\RenewDocumentCommand{\chi}{}{\latexchi\movedownsub}

\newcommand{\calG}{\mathcal{G}}

\newcommand{\calN}{\mathcal{N}}

\newcommand{\mbb}{\bm{b}}

\newcommand{\mbd}{\bm{d}}

\newcommand{\mbh}{\bm{h}}

\newcommand{\mbI}{\bm{I}}

\newcommand{\mbm}{\bm{m}}

\newcommand{\mbp}{\bm{p}}

\newcommand{\mbq}{\bm{q}}

\newcommand{\mbr}{\bm{r}}

\newcommand{\mbW}{\bm{W}}
\newcommand{\mbx}{\bm{x}}

\newcommand{\mby}{\bm{y}}

\DeclareSymbolFont{bsfletters}{OT1}{cmss}{bx}{n}
\DeclareSymbolFont{ssfletters}{OT1}{cmss}{m}{n}
\DeclareMathSymbol{\bsfGamma}{0}{bsfletters}{'000}
\DeclareMathSymbol{\ssfGamma}{0}{ssfletters}{'000}
\DeclareMathSymbol{\bsfDelta}{0}{bsfletters}{'001}
\DeclareMathSymbol{\ssfDelta}{0}{ssfletters}{'001}
\DeclareMathSymbol{\bsfTheta}{0}{bsfletters}{'002}
\DeclareMathSymbol{\ssfTheta}{0}{ssfletters}{'002}
\DeclareMathSymbol{\bsfLambda}{0}{bsfletters}{'003}
\DeclareMathSymbol{\ssfLambda}{0}{ssfletters}{'003}
\DeclareMathSymbol{\bsfXi}{0}{bsfletters}{'004}
\DeclareMathSymbol{\ssfXi}{0}{ssfletters}{'004}
\DeclareMathSymbol{\bsfPi}{0}{bsfletters}{'005}
\DeclareMathSymbol{\ssfPi}{0}{ssfletters}{'005}
\DeclareMathSymbol{\bsfSigma}{0}{bsfletters}{'006}
\DeclareMathSymbol{\ssfSigma}{0}{ssfletters}{'006}
\DeclareMathSymbol{\bsfUpsilon}{0}{bsfletters}{'007}
\DeclareMathSymbol{\ssfUpsilon}{0}{ssfletters}{'007}
\DeclareMathSymbol{\bsfPhi}{0}{bsfletters}{'010}
\DeclareMathSymbol{\ssfPhi}{0}{ssfletters}{'010}
\DeclareMathSymbol{\bsfPsi}{0}{bsfletters}{'011}
\DeclareMathSymbol{\ssfPsi}{0}{ssfletters}{'011}
\DeclareMathSymbol{\bsfOmega}{0}{bsfletters}{'012}
\DeclareMathSymbol{\ssfOmega}{0}{ssfletters}{'012}

\makeatletter
\newcommand*\rel@kern[1]{\kern#1\dimexpr\macc@kerna}
\newcommand*\widebar[1]{%
  \begingroup
  \def\mathaccent##1##2{%
    \rel@kern{0.8}%
    \overline{\rel@kern{-0.8}\macc@nucleus\rel@kern{0.2}}%
    \rel@kern{-0.2}%
  }%
  \macc@depth\@ne
  \let\math@bgroup\@empty \let\math@egroup\macc@set@skewchar
  \mathsurround\z@ \frozen@everymath{\mathgroup\macc@group\relax}%
  \macc@set@skewchar\relax
  \let\mathaccentV\macc@nested@a
  \macc@nested@a\relax111{#1}%
  \endgroup
}
\makeatother

\DeclareMathOperator*{\concat}{\scalerel*{\parallel}{\sum}}

\newcommand{\ifbcdot}[1]{\ifblank{#1}{\cdot}{#1}}

\DeclarePairedDelimiterX\abs[1]{\lvert}{\rvert}{\ifbcdot{#1}}
\DeclarePairedDelimiterX\parens[1]{(}{)}{\ifbcdot{#1}}
\DeclarePairedDelimiterX\brk[1]{[}{]}{\ifbcdot{#1}}
\DeclarePairedDelimiterX\braces[1]{\{}{\}}{\ifbcdot{#1}}
\DeclarePairedDelimiterX\angles[1]{\langle}{\rangle}{\ifblank{#1}{\cdot,\cdot}{#1}}
\DeclarePairedDelimiterX\ip[2]{\langle}{\rangle}{\ifbcdot{#1},\ifbcdot{#2}}
\DeclarePairedDelimiterX\norm[1]{\lVert}{\rVert}{\ifbcdot{#1}}
\DeclarePairedDelimiterX\ceil[1]{\lceil}{\rceil}{\ifbcdot{#1}}
\DeclarePairedDelimiterX\floor[1]{\lfloor}{\rfloor}{\ifbcdot{#1}}

\DeclareFontFamily{U}{matha}{\hyphenchar\font45}
\DeclareFontShape{U}{matha}{m}{n}{
      <5> <6> <7> <8> <9> <10> gen * matha
      <10.95> matha10 <12> <14.4> <17.28> <20.74> <24.88> matha12
      }{}
\DeclareSymbolFont{matha}{U}{matha}{m}{n}
\DeclareFontSubstitution{U}{matha}{m}{n}

\DeclareFontFamily{U}{mathx}{\hyphenchar\font45}
\DeclareFontShape{U}{mathx}{m}{n}{
      <5> <6> <7> <8> <9> <10>
      <10.95> <12> <14.4> <17.28> <20.74> <24.88>
      mathx10
      }{}
\DeclareSymbolFont{mathx}{U}{mathx}{m}{n}
\DeclareFontSubstitution{U}{mathx}{m}{n}

\DeclareMathDelimiter{\vvvert}{0}{matha}{"7E}{mathx}{"17}
\DeclarePairedDelimiterX\vertiii[1]{\vvvert}{\vvvert}{\ifbcdot{#1}}

\DeclarePairedDelimiterXPP\trace[1]{\operatorname{Tr}}{(}{)}{}{\ifbcdot{#1}} 
\DeclarePairedDelimiterXPP\col[1]{\operatorname{col}}{\{}{\}}{}{\ifbcdot{#1}} 
\DeclarePairedDelimiterXPP\row[1]{\operatorname{row}}{\{}{\}}{}{\ifbcdot{#1}} 
\DeclarePairedDelimiterXPP\erf[1]{\operatorname{erf}}{(}{)}{}{\ifbcdot{#1}}
\DeclarePairedDelimiterXPP\erfc[1]{\operatorname{erfc}}{(}{)}{}{\ifbcdot{#1}}
\DeclarePairedDelimiterXPP\KLD[2]{D}{(}{)}{}{\ifbcdot{#1}\, \delimsize\|\, \ifbcdot{#2}} 
\DeclarePairedDelimiterXPP\op[2]{\operatorname{#1}}{(}{)}{}{#2} 

\DeclarePairedDelimiterXPP\indicate[1]{{\bf 1}}{\{}{\}}{}{\ifbcdot{#1}}

\NewDocumentCommand\ofrac{s m}{%
	\IfBooleanTF#1%
	{\dfrac{1}{#2}}%
	{\frac{1}{#2}}%
}
\NewDocumentCommand\ddfrac{s m m}{%
	\IfBooleanTF#1%
	{\dfrac{\mathrm{d} {#2}}{\mathrm{d} {#3}}}%
	{\frac{\mathrm{d} {#2}}{\mathrm{d} {#3}}}%
}
\NewDocumentCommand\ppfrac{s m m}{%
	\IfBooleanTF#1%
	{\dfrac{\partial {#2}}{\partial {#3}}}%
	{\frac{\partial {#2}}{\partial {#3}}}%
}

\providecommand\given{}

\DeclarePairedDelimiterX\Set[2]\{\}{%
\renewcommand\given{\SetSymbol[\delimsize]{#1}}
#2
}
\DeclarePairedDelimiterX\Setc[1]\{\}{%
\renewcommand\given{\SetSymbol{:}}
#1
}

\NewDocumentCommand\set{s o m}{%
	\IfBooleanTF#1%
	{\IfValueTF{#2}{\Set*{#2}{#3}}{\Setc*{#3}}}%
	{\IfValueTF{#2}{\Set{#2}{#3}}{\Setc{#3}}}%
}

\NewDocumentCommand{\evalat}{ s O{\big} m e{_^} }{%
\IfBooleanTF{#1}%
{\left. #3 \right|}{#3#2|}%
\IfValueT{#4}{_{#4}}%
\IfValueT{#5}{^{#5}}%
}

\providecommand\given{}
\DeclarePairedDelimiterXPP\cprob[1]{}(){}{
\renewcommand\given{\nonscript\,\delimsize\vert\allowbreak\nonscript\,\mathopen{}}%
#1%
}
\DeclarePairedDelimiterXPP\cexp[1]{}[]{}{
\renewcommand\given{\nonscript\,\delimsize\vert\allowbreak\nonscript\,\mathopen{}}%
#1%
}

\DeclareDocumentCommand \P { s e{_^} d() g } {%
	\mathbb{P}%
	\IfBooleanTF{#1}%
		{
			\IfValueT{#2}{_{#2}}%
			\IfValueT{#3}{^{#3}}%
			\IfValueTF{#5}{\cprob{#4 \given #5}}{\IfValueT{#4}{\cprob{#4}}}%
		}%
		{
			\IfValueT{#2}{_{#2}}%
			\IfValueT{#3}{^{#3}}%
			\IfValueTF{#5}{\cprob*{#4 \given #5}}{\IfValueT{#4}{\cprob*{#4}}}%
		}%
}

\DeclareDocumentCommand \E { s e{_^} o g } {%
	\mathbb{E}%
	\IfBooleanTF{#1}%
		{
			\IfValueT{#2}{_{#2}}%
			\IfValueT{#3}{^{#3}}%
			\IfValueTF{#5}{\cexp{#4 \given #5}}{\IfValueT{#4}{\cexp{#4}}}%
		}%
		{
			\IfValueT{#2}{_{#2}}%
			\IfValueT{#3}{^{#3}}%
			\IfValueTF{#5}{\cexp*{#4 \given #5}}{\IfValueT{#4}{\cexp*{#4}}}%
		}%
}

\ExplSyntaxOn
\NewDocumentCommand \dist {m o o} {%
\mathrm{#1}\left(%
	\IfValueT{#3}{%
		\tl_if_blank:nTF{ #3 }{\cdot\, \middle|\, }{#3\, \middle|\, }%
	}
	\IfValueT{#2}{#2}%
\right)%
}
\ExplSyntaxOff

\NewDocumentCommand {\cbrace} {t+ D[]{black} D(){\widthof{#5}} m m } {%
	\begingroup%
		\color{#2}
		\IfBooleanTF{#1}{%
			\overbrace{#4}^%
		}{
			\underbrace{#4}_%
		}%
		{\parbox[c]{#3}{\centering\footnotesize{#5}}}%
	\endgroup%
}

\let\oldforall\forall
\renewcommand{\forall}{\oldforall \, }

\let\oldexist\exists
\renewcommand{\exists}{\oldexist \, }

\graphicspath{{./Figures/}{./figures/}}
\DeclareDocumentCommand{\includeCroppedPdf}{ o O{./Figures/} m }{
	\IfFileExists{#2#3-crop.pdf}{}{%
		\immediate\write18{pdfcrop #2#3.pdf #2#3-crop.pdf}}%
	\includegraphics[#1]{#2#3-crop.pdf}
}

\makeatletter
\newcommand*{\addFileDependency}[1]{
  \typeout{(#1)}
  \@addtofilelist{#1}
  \IfFileExists{#1}{}{\typeout{No file #1.}}
}
\makeatother

\definecolor{gray90}{gray}{0.9}

\ifx\nohighlights\undefined

	\newcommand{\msout}[1]{\text{\color{green} \sout{\ensuremath{#1}}}}
	\newcommand{\del}[1]{{\color{green}\ifmmode \msout{#1}\else\sout{#1}\fi}}
\else

	\newcommand{\msout}[1]{#1}
	\newcommand{\del}[1]{#1}
\fi

\newcommand{\hhide}[1]{}

\ifx\diagnoselabel\undefined
	\relax
\else
	\makeatletter
	\def\@testdef #1#2#3{%
		\def\reserved@a{#3}\expandafter \ifx \csname #1@#2\endcsname
			\reserved@a  \else
			\typeout{^^Jlabel #2 changed:^^J%
				\meaning\reserved@a^^J%
				\expandafter\meaning\csname #1@#2\endcsname^^J}%
			\@tempswatrue \fi}
	\makeatother
\fi
